\documentclass[lettersize,journal]{IEEEtran}
\usepackage{amsmath,amsfonts}
\usepackage{algorithmic}
\usepackage[linesnumbered,ruled,vlined]{algorithm2e}
\usepackage{array}
\usepackage[caption=false,font=normalsize,labelfont=sf,textfont=sf]{subfig}
\usepackage{textcomp}
\usepackage{stfloats}
\usepackage{url}
\usepackage{verbatim}
\usepackage{graphicx}
\usepackage{makecell}
\usepackage{booktabs} 
\usepackage{multirow}
\usepackage{threeparttable}
\usepackage{cite}
\usepackage{blindtext}
\usepackage{placeins}
\usepackage[colorlinks,
linkcolor=blue,
anchorcolor=blue,
citecolor=blue
]{hyperref}

\begin{document}

\title{Contact-Agnostic Action Strategy for Two-Fingered
Reorientation with External Forces}

\author{Qiyin Huang, Yijia Lei, Yue Wang, Lunwei Zhang,
Yen Hang Zhou, Tiemin Li, and Yao Jiang%
\thanks{All authors are with Tsinghua University, Beijing, China.}%
\thanks{Corresponding author: Yao Jiang.}%
}

\maketitle

\begin{abstract} 
Two-fingered in-hand object reorientation showcases robotic dexterity but requires real-time adaptation to dynamically changing external forces, such as gravity and contact forces. Among them, object-environment contact forces are particularly complex and variable. Existing methods often rely on explicit contact modeling or task-specific pre-training, limiting their applicability across diverse conditions. To address this, we propose a contact-agnostic reorientation strategy that operates without prior knowledge of object-environment contact mechanics. We introduce a novel action evaluation metric grounded in task-level progress and perceptual expectation, bypassing the need for explicit contact modeling. By combining this metric with a local linear approximation and a stepwise reward update mechanism, our method enables online action adjustment in unfamiliar scenarios. Experimental results demonstrate that the proposed approach consistently outperforms baseline methods across a variety of objects and contact states, without requiring pre-training. The successful reorientation of a cabled charger under external disturbances further validates the approach's real-world applicability.
\end{abstract}

\def\abstractname{Note to Practitioners}
\begin{abstract}
This paper addresses the common industrial challenge of in-hand object reorientation, where a robot must adjust its grasp on a part—for example, aligning a bolt for insertion or orienting a tool for use—without setting it down. Traditionally, this requires expensive multi-fingered hands, precise environmental modeling, or rigid regrip stations. We present a strategy that enables standard two-finger grippers to reorient objects by actively leveraging external forces, such as gravity or contact with a work surface. By integrating visual and tactile sensors, the system evaluates the success of its movements in real-time and adjusts its strategy on the fly. This approach eliminates the need for time-consuming pre-training or complex physics models, significantly improving flexibility and reducing setup time for new product lines. The system is shown to handle diverse objects, including those with flexible cables, making it highly reliable for unstructured environments like bin picking or logistics. Future work will focus on extending this sensory-driven approach to other contact-rich tasks, such as precision insertion and cutting.
\end{abstract}


\section{Introduction}
In-hand object reorientation is a fundamental yet challenging task for dexterous robotic manipulation, especially for grippers with limited degrees of freedom \cite{SUOMALAINEN2022104224}. This task requires the robot to accurately adjust the object's orientation in the presence of uncertain external forces\cite{tao2023Visual, peng2024Learning}. 
External forces refer to forces other than the forces applied by the finger contact \cite{nikhilchavan2014Extrinsic}, including gravity, environmental contact force, etc. These forces further complicate the reorientation task by inducing unintended object motion, destabilizing contact configurations, and increasing uncertainty in action outcomes. 

Adapting to external forces is a key enabler of dexterous in-hand manipulation. In tasks such as assembly, contact with neighboring components may either assist or hinder alignment; in object retrieval from a container, contact forces from container walls are often leveraged to facilitate reorientation while avoiding collisions with the container opening.
This role of external forces becomes especially important for low-DOF grippers, whose limited degrees of freedom constrain their ability to generate internal object motion. In such cases, external contacts are not just disturbances to avoid—they provide the additional constraints and forces needed to initiate and guide movement. Therefore, the ability to adapt to external forces is essential to expanding the manipulation capabilities of low-DOF grippers.

Most existing approaches address external forces in object reorientation by relying on either pre-defined contact models or scenario-specific pre-training. Although these methods can be effective in familiar environments, they often struggle to generalize to new or changing environments \cite{jinda2021Nextgeneration}. This limitation stems from the inherently complex nature of external contacts—characterized by sudden changes, discontinuities, and nonlinear behaviors—which makes it challenging to construct accurate contact models in advance or adapt them online.

To address the challenges of complex external forces and intractable contact modeling, a contact-agnostic strategy is proposed that integrates perceptual outcome estimation with online action adjustment. The core insight is to reframe the intractable global optimization problem as a local decision-making process guided by perception. Unlike global modeling, which strives to construct a comprehensive value function across the entire state-action space—a process often plagued by the curse of dimensionality and contact discontinuities—the proposed approach focuses on estimating the gradient of change within the immediate state neighborhood. This strategy significantly reduces the volume of action exploration required, allowing the system to adapt online to varying external forces. Specifically, a multi-modal perceptual expectation is constructed to serve as a task-relevant reference, independent of specific environmental contact mechanics. On this basis, a local feedback mechanism continuously refines actions based on real-time deviations. This synergy enables goal-driven evaluation and efficient adaptation without relying on task-specific pre-training or explicit contact priors.

In summary, the contributions are:
\begin{itemize}
    \item Action evaluation via perception expectations: A visual-tactile framework is designed to establish task-relevant references, enabling consistent action evaluation across diverse and unknown contact conditions.
    \item Efficient action adjustment via local estimation: By leveraging gradient-based exploration and stepwise reward updates, this strategy allows the robot to dynamically assess action efficacy and adjust control inputs in real-time, bypassing the computational burden of global modeling.
    \item Adaptive performance across diverse scenarios: Without pre-training or contact-specific modeling, the proposed method achieves performance comparable or superior to trained baselines. It demonstrates strong stability under sudden disturbances and nonlinear contact dynamics in unstructured settings.
\end{itemize}

\section{Related Work}
\subsection{Two-fingered Object Reorientation}
In-hand object reorientation is a fundamental capability in dexterous manipulation, enabling tasks such as tool usage and precision assembly \cite{sudharshan2024NeuralFeels}. While multi-fingered hands or coordinated arm-hand systems can achieve complex reorientations through internal actuation alone \cite{guotao2024LinkageDriven}, recent studies have shown that even low-DOF grippers can perform in-hand reorientation by exploiting external forces—such as gravity and environmental contacts—as auxiliary sources of manipulation \cite{marco2020TwoFingered, yuki2024Robust}.

These approaches reveal a critical distinction: the reorientation capability is not merely determined by the gripper's internal actuation, but can be functionally extended by interacting with external forces \cite{nikhilchavan2014Extrinsic}. This is particularly important in the context of two-finger grippers, whose limited degrees of freedom render many reorientation tasks infeasible without assistance \cite{aude2019Trends}. In this setting, external forces are no longer optional aids, but essential components of the manipulation process.

However, leveraging such external interactions introduces a dual challenge: while external forces enable new manipulation possibilities, they also introduce uncertainty, variability, and non-repeatability into the contact outcomes. Existing works largely follow two directions to address this issue: (1) Constraint-modeling approaches, which explicitly formulate contact constraints and motion feasibility using analytical or physics-based models; and (2) task-specific pre-training approaches, which learn manipulation strategies directly from interaction data, thereby circumventing the need for accurate physical modeling.

Methods based on contact constraint modeling explicitly define the feasible action space by modeling the force boundaries within the finger--object--environment contact chain. Representative models include friction cones \cite{jingyi20206DFC, azad2021polyhedral} and limit surface models \cite{nicholas1999Modeling}. Such contact constraints enable the robot to efficiently evaluate and select feasible actions. For example, by constructing rigid point contact models \cite{chao2021Status, neel2022Manipulation} or using limit surface representations \cite{nikhil2020Planar}, researchers have derived motion cones to facilitate fast and reliable action planning. However, real-world manipulation often involves multi-point, compliant, or shifting contacts \cite{quentin2024Contact}, which introduce discontinuities and uncertainty into the contact model. 

Task-specific pre-training approaches, exemplified by deep reinforcement learning (DRL), aim to bypass explicit contact modeling by directly learning state-to-action mappings from interaction data \cite{athanasioss.2017Survey}. However, the high-dimensional and continuous nature of the state space, coupled with complex and often unpredictable force interactions, typically necessitates large-scale data collection in simulation for training \cite{inigo2023Review}. This reliance on simulated data often leads to biased sampling, which can result in overfitting and poor generalization to novel scenarios \cite{nahavandi2024machine}. 
To address the above challenges, some researchers have explored introducing more efficient representations of input information \cite{entong2024Sim2Real, chen2020SwingBot}. Others have sought to improve generalization by decomposing the task into hierarchical or modular sub-tasks  \cite{xianyi2024Enhancing, xiang2023Learning}. Experimental results demonstrate that both input representation and task decomposition can enable data-efficient training while maintaining generalization performance.
However, in the context of in-hand reorientation, existing methods still largely depend on prior learning of object properties or contact models. This reliance on pre-learned information limits the adaptability of these approaches when faced with novel objects or previously unseen contact conditions.

A notable recent work is TEXterity \cite{antonia2024TEXterity}, which presents a comprehensive framework for exploiting environmental contacts for object reorientation. However, its formulation assumes access to task-specific priors, including a pre-acquired object point cloud, the geometry and orientation of environmental support surfaces, and predefined point-contact interaction models. The proposed approach, by contrast, relies solely on onboard sensing without requiring such geometric or contact-mode priors. As a result, TEXterity targets a complementary problem regime, making a direct baseline comparison not entirely appropriate for the current evaluation setting.

In response to these limitations and development, a contact-agnostic action evaluation metric combined with an online action adjustment mechanism is proposed. This approach enables real-time adaptation without relying on predefined contact models or extensive pre-training, thereby enhancing the robot's ability to handle a wide range of external contacts during in-hand reorientation.

\subsection{Online Optimization in Manipulation}

Online optimization enables real-time adaptation during robotic interaction, which is particularly critical for contact-rich and dynamic manipulation tasks \cite{sebastienIntroduction}. However, their applicability to in-hand reorientation remains limited.

Model Predictive Control (MPC) relies on accurate dynamic models to solve constrained optimization problems at each timestep. It has been applied to robotic manipulation for trajectory tracking and force control \cite{max2021Review, tobias2023Model}. However, modeling of discontinuous or nonlinear contacts--common in in-hand reorientation--is notoriously difficult \cite{lukas2020LearningBased}. Moreover, the computational burden of solving constrained optimization problems at high frequency makes real-time deployment challenging in the face of multiple contact modes and high-dimensional action spaces \cite{ugo2018DataDriven}.

Online reinforcement learning (RL) update policies through continuous interaction \cite{han2023Adaptive} and have been applied to manipulation tasks involving contact uncertainty. Yet, due to the limited interaction episodes and high-dimensionality of state-action spaces of in-hand manipulation, online RL typically suffers from poor sample efficiency and slow convergence \cite{sinclair2023adaptive, wagenmaker2023leveraging}. These limitations often lead to slow convergence and suboptimal real-time behavior.

Although existing online optimization methods have made progress in some manipulation tasks, they often become data-inefficient or even fail in contact-rich environments due to two core limitations: 
(1) Unreliable action evaluation, often stemming from reliance on contact-specific models;
(2) Limited real-time performance, as nonlinear optimization and high-dimensionality make real-time updates computationally expensive.
To address these issues, a contact-agnostic action adjustment strategy is proposed, which enables online adaptability without relying on predefined models. Furthermore, a local approximation of action-induced motion is exploited to reduce the effective complexity of action evaluation, facilitating efficient online
updates without requiring a global contact-dynamics model.

\section{Problem and Framework}

\subsection{Problem}

The contact-rich reorientation task requires the gripper to manipulate an object through continuous interactions with the environment so that the object reaches a desired orientation.

The manipulation considered in this work is restricted to a fixed task plane $\Pi$. Let
$\boldsymbol{B}=[\boldsymbol{b}_1,\boldsymbol{b}_2]\in\mathbb{R}^{3\times 2}$
denote an orthonormal basis of $\Pi$, satisfying
$\boldsymbol{B}^{\mathsf T}\boldsymbol{B}=\boldsymbol{I}_2$, and let
$\boldsymbol{n}_{\Pi}=\boldsymbol{b}_1\times\boldsymbol{b}_2$
denote its unit normal. Although the physical object pose is described by
$\boldsymbol{P}_o\in\mathbb{R}^3$ and $\boldsymbol{R}_o\in SO(3)$,
the task-relevant state is represented in the coordinates of $\Pi$ by the
planar translation $\boldsymbol{d}\in\mathbb{R}^2$ and the signed rotation
angle $\theta\in\mathbb{R}$ about $\boldsymbol{n}_{\Pi}$.

Formally, the state space $\mathcal{S}$ of the object and action space $\mathcal{A}$ of the gripper are defined as follows.
\begin{equation}
    \mathcal{S}
    =
    \left\{
    \boldsymbol{s}
    \mid
    \boldsymbol{s}=(\boldsymbol{d},\theta)
    \right\},
    \label{equ::object state}
\end{equation}
\begin{equation}
    \mathcal{A}
    =
    \left\{
    \boldsymbol{a}
    \mid
    \boldsymbol{a}
    =
    (\Delta\boldsymbol{p}_h,\Delta\theta_h,\Delta F_h)
    \right\},
    \label{equ::action space}
\end{equation}
where $\boldsymbol{d}$ denotes the object translation expressed in the
task-plane basis $\boldsymbol{B}$, and $\theta$ denotes the signed object
rotation about $\boldsymbol{n}_{\Pi}$.
The gripper translation
$\Delta\boldsymbol{p}_h\in\mathbb{R}^2$ is expressed in the same task-plane
coordinates, $\Delta\theta_h\in\mathbb{R}$ denotes the incremental gripper
rotation about $\boldsymbol{n}_{\Pi}$, and $\Delta F_h\in\mathbb{R}$ denotes
the change in grasping force.
The proposed framework is in principle applicable to all three control inputs.
In most experiments, only translational motion and grasping force
$(\Delta\boldsymbol{p}_h,\Delta F_h)$ are activated, which already enable reliable reorientation under contact constraints.
For the demonstration, the rotational component $\Delta\theta_h$ is added to the candidate action set, enabling coupled translational--rotational control.

The objective of the task is to minimize the error between $\boldsymbol{s}_t$ and the target state $\boldsymbol{s}^*$, by selecting a sequence of actions that gradually drive the system toward the goal.

\begin{figure}[t]
    \centering
    \includegraphics[width = 1.0\linewidth]{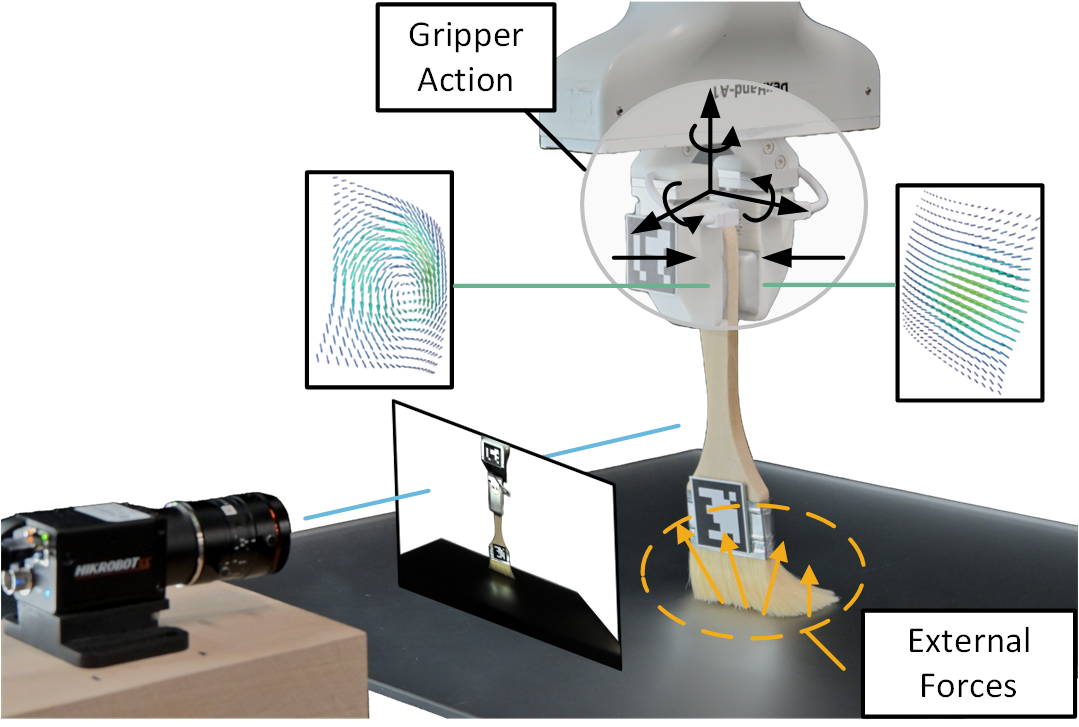}
    \caption{Two-fingered reorientation system.}
    \label{fig::two-fingered reorientation}
\end{figure}

\begin{table}[t]
    \centering
    \caption{List of Notations.}
    \label{tab::parameters}
    \begin{tabular}{p{0.25\columnwidth}p{0.65\columnwidth}}
        \toprule
        Symbol & Description \\
        \midrule
        \multicolumn{2}{l}{\textit{Sets and Frames}} \\
        $\mathcal{S}, \mathcal{A}$ & State space and Action space \\
        $\Pi, \boldsymbol{B}, \boldsymbol{n}_{\Pi}$ & Task plane, its orthonormal basis, and unit normal \\
        \midrule
        \multicolumn{2}{l}{\textit{State and Action}} \\
        $\boldsymbol{s}, \boldsymbol{s}^*$ & Current object state and target state \\
        $\boldsymbol{P}_o, \boldsymbol{R}_o$ & Physical object position and orientation \\
        $\boldsymbol{d}, \theta$ & Planar object translation and signed rotation angle \\
        $\boldsymbol{a}$ & Gripper action vector \\
        $\Delta\boldsymbol{p}_h, \Delta\theta_h, \Delta F_h$ & Incremental planar translation, rotation, and grasping-force change \\
        $\hat{l}$ & Normalized action amplitude \\
        \midrule
        \multicolumn{2}{l}{\textit{Kinematics and Perception}} \\
        $\boldsymbol{\delta}_j$ & Displacement vector of tactile marker $j$ \\
        $o, \mathbb{E}[o]$ & Perceptual observation and expectation \\
        $e_0, \boldsymbol{u}_0$ & Instantaneous sub-targets for rotation and planar translation \\
        $J_\theta, \boldsymbol{J}_d$ & Local Jacobians for rotation and planar translation \\
        \midrule
        \multicolumn{2}{l}{\textit{Utility and Optimization}} \\
        $G(\cdot)$ & Action evaluation metric \\
        $R(\cdot)$ & Total action reward function \\
        $U_\mathrm{n}, U_\mathrm{f}, U_\mathrm{e}$ & Immediate, Future, and Uncertainty utilities \\
        $\beta_1, \beta_2$ & Weighting coefficients for utilities \\
        $N_{\boldsymbol{a}}$ & Execution count of action $\boldsymbol{a}$ \\
        $\alpha$ & Update rate for utility estimation \\
        \bottomrule
    \end{tabular}
\end{table}

\subsection{Framework}

Due to the diversity and uncertainty of contact dynamics, the outcome of a given action $\boldsymbol{a}_t$ under state $\boldsymbol{s}_t$ is often difficult to predict, which complicates the formulation of a unified strategy for action evaluation and decision-making. Two primary challenges arise.

First, varying contact modalities—such as point, surface, sliding, and rolling contact—produce nonlinear and time-varying effects on action outcomes, preventing the establishment of a generalizable evaluation standard. Consequently, contact-model-dependent methods often fail to adapt to novel objects or unseen contact scenarios, limiting policy reusability.

Second, even with a unified evaluation criterion, action outcomes remain highly sensitive to both instantaneous contact states and their temporal evolution. This state- and time-dependent variability undermines the stability and predictability of action selection, especially in dynamically changing contact environments.

In summary, this work addresses: (1) the absence of a contact-agnostic action evaluation, and (2) the difficulty of predicting action outcomes due to their strong dependence on contact states. To address these challenges, the proposed framework consists of two complementary components. Section IV introduces a contact-agnostic action evaluation metric based on task progress and perceptual expectations, providing a unified basis for evaluating actions under different contact conditions. Section V develops an online action adjustment strategy that incorporates local reward estimation, future utility, and uncertainty utility through stepwise updates, enabling real-time adaptation to changing contact conditions.

\begin{figure*}[t]
    \centering
    \includegraphics[width = 1.0\linewidth]{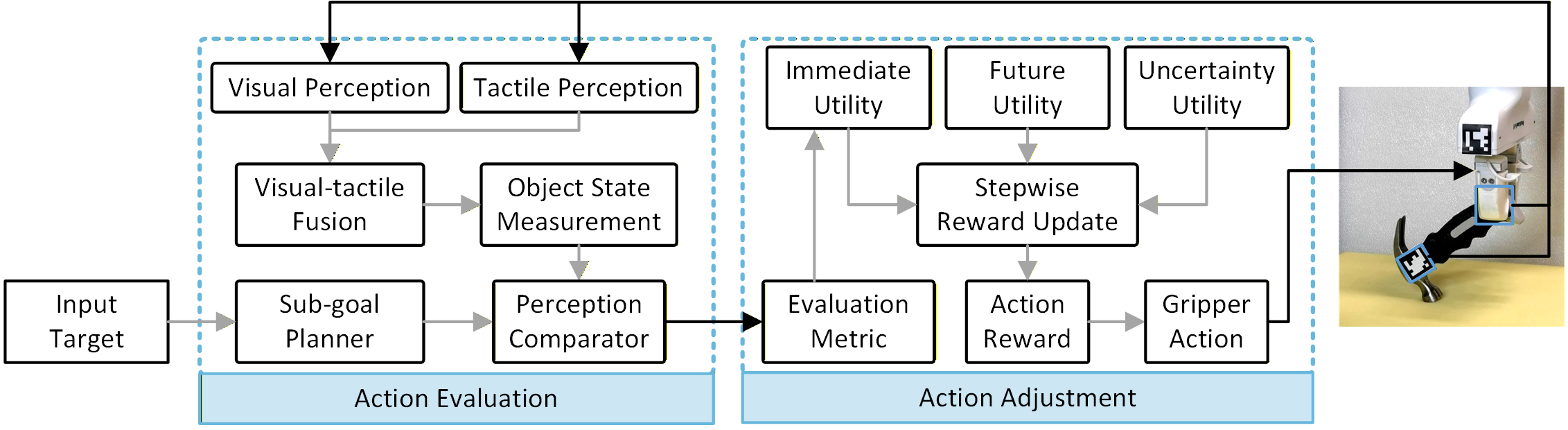}
    \caption{Overview of the proposed adaptive reorientation framework. The evaluation module (left) provides a generalizable action assessment by comparing the current state with dynamically determined sub-goals using visual and tactile perception, without relying on explicit contact modeling. It ensures accurate differentiation between actions of varying quality to support reliable decision-making. The action adjustment module (right) updates actions based on evaluation results through four key mechanisms: (1) local continuous evaluation leveraging state differentials, (2) an uncertainty utility to address temporal decay effects, (3) incorporation of future utility to account for the influence of actions on subsequent states, and (4) a unified utility update and prediction process that combines immediate, future, and uncertainty utilities. Together, these components enable stable, efficient, and real-time action selection under unknown and evolving contact conditions.}
    \label{fig::framework}
\end{figure*}

A schematic of the two--finger reorientation system is shown in Fig. \ref{fig::two-fingered reorientation}. The notations used throughout this paper are summarized in Table \ref{tab::parameters}. And the proposed real-time action evaluation and adjustment framework is illustrated in Fig. \ref{fig::framework}.

It is important to distinguish between immediate utility and action reward. Immediate utility refers to the direct evaluation of a single action, reflecting its immediate effect on task progression. In contrast, the action reward integrates immediate utility with additional components—such as exploration incentives and inter-action coupling—thereby supporting strategy updates and enabling more informed decision-making.

\subsection{Assumptions}

This work focuses on how a two-finger gripper, without increasing its actuation complexity, can actively adapt to and make use of unknown and dynamically changing external forces, thereby expanding its reorientation capability beyond its intrinsic mechanical limits.

The gripper action space is assumed to contain feasible motions for object reorientation. In extreme cases, such as full object constraint or severely limited movement, external assistance or additional planning may be necessary.

The mapping from action amplitude $\hat{l}$ to the induced object motion--rotation $\Delta\theta$ and planar translation $\Delta\boldsymbol{d}$--is assumed to be smooth within a local neighborhood of $\hat{l}=0$.

To enhance computational efficiency and real-time performance, this study adopts an incremental optimization strategy that progressively guides the object toward the target posture, rather than solving for a globally optimal action at each step.

\hypertarget{sec::action evaluation}{}

\section{Action Evaluation}

This section analyzes the kinematics of the object during the reorientation process with a two-fingered gripper, proposing an action evaluation method that is independent of the form of specific external force forms. To improve evaluation accuracy, a variable-weight fusion of visual and tactile inputs is introduced to leverage their complementary characteristics.

\hypertarget{sec::kinematic analysis}{}
\subsection{Kinematic Analysis of Two-fingered Reorientation}

\begin{figure}[t]
    \centering
    \includegraphics[width = 0.8\linewidth]{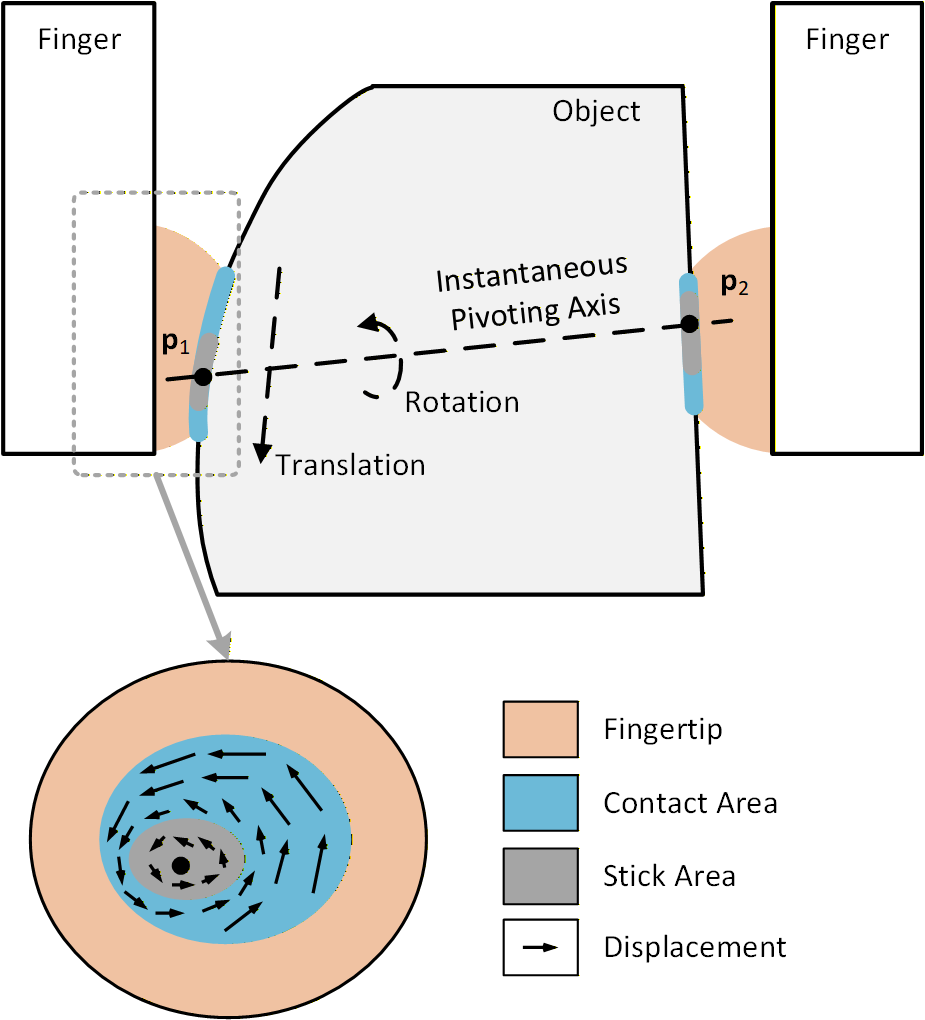}
    \caption{Kinematic characteristics of two-finger reorientation.}
    \label{fig::kinematic analysis}
\end{figure}

The primary goal of kinematic analysis is to bridge the gap between high-dimensional sensory feedback and action evaluation. By extracting essential motion features, this analysis reduces the dimensionality of the feedback, rendering the evaluation process more physically interpretable and computationally efficient.

In two-fingered manipulation with soft grippers, the finger-object interface is characterized by surface contact rather than point contact. Consequently, the contact patch may exhibit non-uniform slip conditions. As illustrated in Fig. \ref{fig::kinematic analysis}, the contact area can be partitioned into a \textit{stick area}, where no relative motion occurs, and a \textit{slip area}, where local sliding exists.

To maintain a stable grasp during reorientation, a sticking area must be preserved within the contact interface. If we consider two points, $\boldsymbol{\mathrm{p}}_1$ and $\boldsymbol{\mathrm{p}}_2$, located within the stick areas of the opposing fingers, their relative velocities with respect to the fingers are zero. Consequently, the instantaneous motion of the object is constrained to a rotation around the axis connecting these two points. This specific motion pattern is defined as \textit{pivoting rotation}.

Thus, provided that partial sticking is maintained, the object's kinematics are constrained to a single degree of freedom: rotation about the axis connecting the stick centers on the two fingers. Let $\boldsymbol{\mathrm{p}}_1$ and $\boldsymbol{\mathrm{p}}_2$ denote the positions of these stick centers in the object frame. The instantaneous angular velocity vector $\boldsymbol{\omega}$ is given by:

\begin{equation}
    \label{equ::object motion with no macro-slip}
    \boldsymbol{\omega}= \omega \cdot \frac{\mathbf{p}_{2}-\mathbf{p}_{1}}{\|\mathbf{p}_{2}-\mathbf{p}_{1}\|}.
\end{equation}

For the planar task considered in this work, the pivot-axis direction coincides with the task-plane normal $\boldsymbol{n}_{\Pi}$. Leveraging these kinematic constraints and the planar task representation defined in Equation (\ref{equ::object state}), the object state is written as:
\begin{equation}
    \boldsymbol{s} = (\boldsymbol{d}, \theta),
    \label{equ::object state of pivoting}
\end{equation}
where $\boldsymbol{d} \in \mathbb{R}^2$ denotes the object translation expressed in the task-plane basis $\boldsymbol{B}$, and $\theta \in \mathbb{R}$ denotes the signed rotation angle about the task-plane normal $\boldsymbol{n}_{\Pi}$.

\subsection{Task-oriented Action Evaluation}

To unify action evaluation across diverse contact conditions, the concept of \textit{perceptual expectation} is introduced, which translates the task goal into the sensory space. Let $g(\cdot)$ denote the mapping from the state space to the fused visual-tactile perceptual space. The perceptual expectation corresponding to the target state $\boldsymbol{s}^*$ is defined as:
\begin{equation}
    \mathbb{E} [o] = g(\boldsymbol{s}^*),
    \label{equ::perception expectation (v1)}
\end{equation}
Similarly, the current observation is given by $o_i = g(\hat{\boldsymbol{s}}_i)$. This mapping $g(\cdot)$ integrates global geometric cues from vision and local contact information from tactile sensing, ensuring reliable state estimation under partial observability and contact uncertainty.

To quantify the contribution of an action $\boldsymbol{a}_i$ to the task, we assess the alignment between the predicted post-action observation $\hat{o}_{i+1|\boldsymbol{s}_i,\boldsymbol{a}_i}$ and the perceptual expectation. A general \textit{similarity measure} is defined as:
\begin{equation}
    \mathrm{Sim}: (\hat{o}_{i+1|\boldsymbol{s}_i,\boldsymbol{a}_i}, \mathbb{E} [o]) \mapsto \mathbb{R}.
\end{equation}

Consequently, the action evaluation metric is formulated as:
\begin{equation}
    G(\boldsymbol{s}_i,\boldsymbol{a}_i) = \big( \mathrm{Sim}(\hat{o}_{i+1|\boldsymbol{s}_i,\boldsymbol{a}_i}, \mathbb{E} [o]) \big).
    \label{equ::action evaluation based on similarity}
\end{equation}
Actions driving the system closer to the perceptual expectation yield higher scores. This formulation establishes a contact-agnostic evaluation basis relying solely on perceptual feedback.

\subsection{Tactile Perception}


According to Equation (\ref{equ::action evaluation based on similarity}), the evaluation metric $G(\boldsymbol{s}_i, \boldsymbol{a}_i)$ relies on the perceptual observation $o_i$ and expectation $\mathbb{E} [o]$, which are functions of the object's planar translation $\boldsymbol{d}$ and rotation $\theta$ (Equation (\ref{equ::object state of pivoting})). Therefore, accurate estimation of $\boldsymbol{d}$ and $\theta$ is critical.

Unlike vision, which provides global pose estimates, tactile perception infers object motion from local contact deformations. In this work, landmark-based tactile sensors Tac3D \cite{zhang2022tac3d} are leveraged, whose surface markers provide dense measurements of contact deformation. The sensor outputs the 3D displacement field of tracked landmarks, which is used as the tactile input to our model.

To address this, a \textit{no-slip approximation} is introduced, assuming the entire contact patch adheres to the object. This allows all marker displacements to contribute to motion estimation. Although this approximation introduces minor bias due to micro-slips, it enables robust real-time estimation using the geometric center of the contact patch as the rotation center. This simplification is supported by limit surface theory \cite{nicholas1999Modeling} and prior studies \cite{amin2016Development}, which suggest that the instantaneous rotation center converges toward the geometric center when tangential forces are suppressed.

According to the kinematic analysis (\hyperlink{sec::kinematic analysis}{\uppercase\expandafter{\romannumeral4}-A}), the contact area consists of stick and slip area. While only the stick region accurately reflects object motion, its area diminishes during rotation, reducing data availability. 
To address this, a pragmatic approximation is introduced: the finger is considered to remain fully adhered to the object. This allows all marker displacements to contribute to motion estimation. Although this approximation may cause minor bias due to inevitable micro-slips, it provides a feasible and effective simplification for real-time estimation. The influence of this approximation is analyzed and partially compensated in Section \hyperlink{sec::visual-tactile fusion}{\uppercase\expandafter{\romannumeral4}-D}. 

\begin{figure*}[t]
    \centering
    \includegraphics[width = 0.94\linewidth]{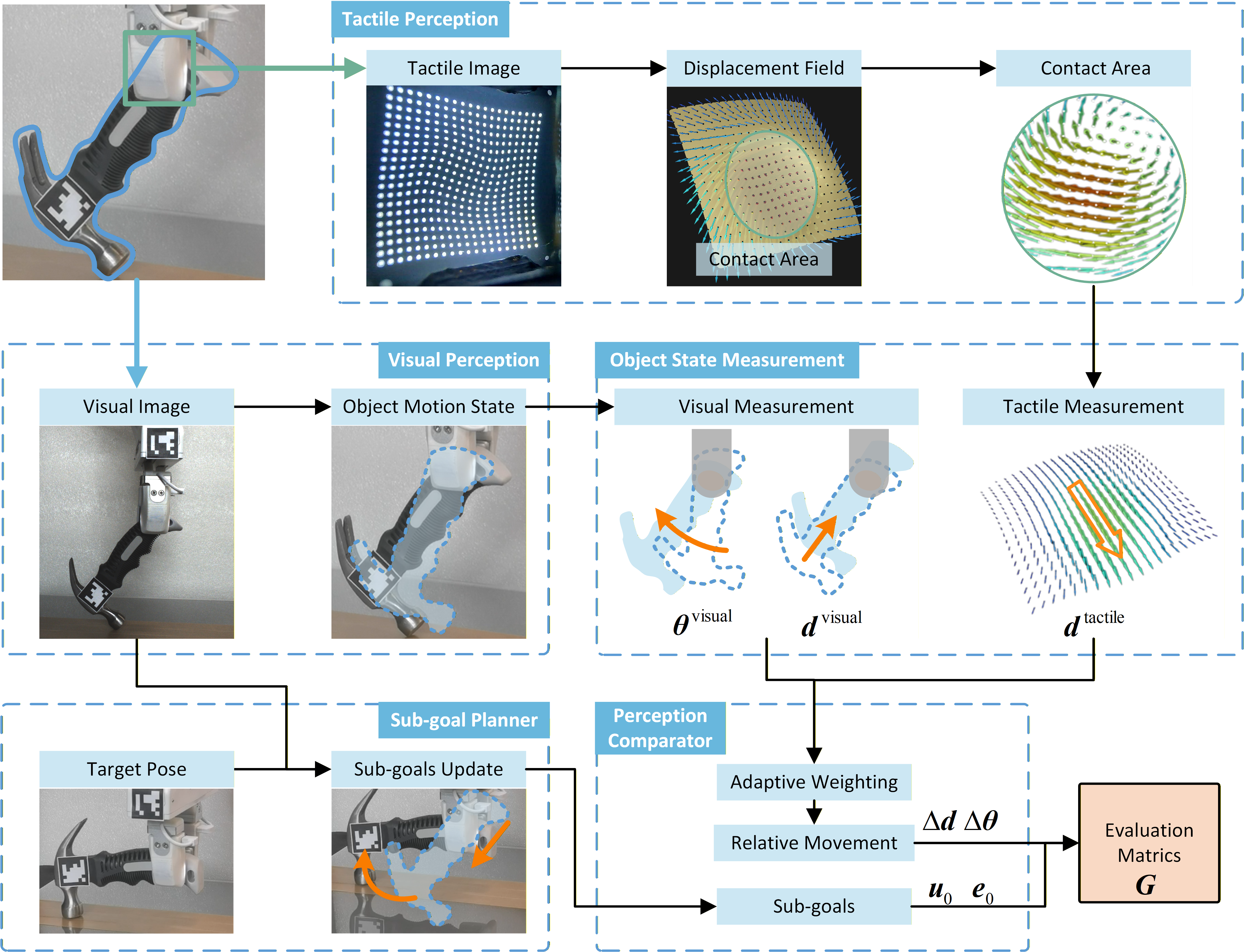}
    \caption{ Action evaluation pipeline. After the robot executes an action, the object pose is captured via visual and tactile sensing. A motion sub-goal is extracted from the visual feedback. The evaluation metric $G$ is then computed by comparing the actual object motion with the sub-goal.}
    \label{fig::tactile measurement}
\end{figure*}

Under this approximation, the object's rotation $\theta^\mathrm{tactile}$ and planar translation $\boldsymbol{d}^\mathrm{tactile}$ are estimated by averaging the rotational and linear displacements of the markers:
\begin{align}
    \label{equ::contact state from tactile feedback (rotation)}
    \theta^\mathrm{tactile}(\boldsymbol{s}_i)
    &\approx
    \frac{
        \displaystyle\sum_{j=1}^{N}
        \left\langle
        \boldsymbol{\theta}_j \times \boldsymbol{\delta}_j,
        \boldsymbol{n}_{\Pi}
        \right\rangle
    }{
        \displaystyle\sum_{j=1}^{N}
        \left\|
        \boldsymbol{B}^{\mathsf T}\boldsymbol{\theta}_j
        \right\|_2^2
    },\\
    \boldsymbol{d}^\mathrm{tactile}(\boldsymbol{s}_i)
    &= \frac{1}{N} \sum\limits_{j=1}^N
    \boldsymbol{B}^{\mathsf T}
    \boldsymbol{\delta}_{j}(\boldsymbol{s}_i),
    \label{equ::contact state from tactile feedback}
\end{align}
where $\boldsymbol{\delta}_{j}$ is the displacement vector of the $j$-th marker, $\boldsymbol{\theta}_{j}$ is its position relative to the contact center, $\boldsymbol{n}_{\Pi}$ is the task-plane normal aligned with the pivot axis direction, and $N$ is the number of markers. The measurement pipeline is illustrated in Fig. \ref{fig::tactile measurement}.

In this paper, the rotation center within the contact area is approximated by the geometric center. This approximation is supported by prior work showing that: (1) when tangential forces are suppressed, the instantaneous rotation center predicted by the limit surface model converges toward the geometric center \cite{nicholas1999Modeling}; and (2) the small deviation between the true and geometric centers has negligible influence on contact forces and moments \cite{amin2016Development}. Accordingly, the geometric center is adopted as an efficient and physically justified approximation. The tactile measurement process is illustrated in Fig. \ref{fig::tactile measurement}.

\hypertarget{sec::visual-tactile fusion}{}
\subsection{Visual-tactile Fusion for Movement Perception}
\begin{figure}[t]
    \centering
    \includegraphics[width = 1.0\linewidth]{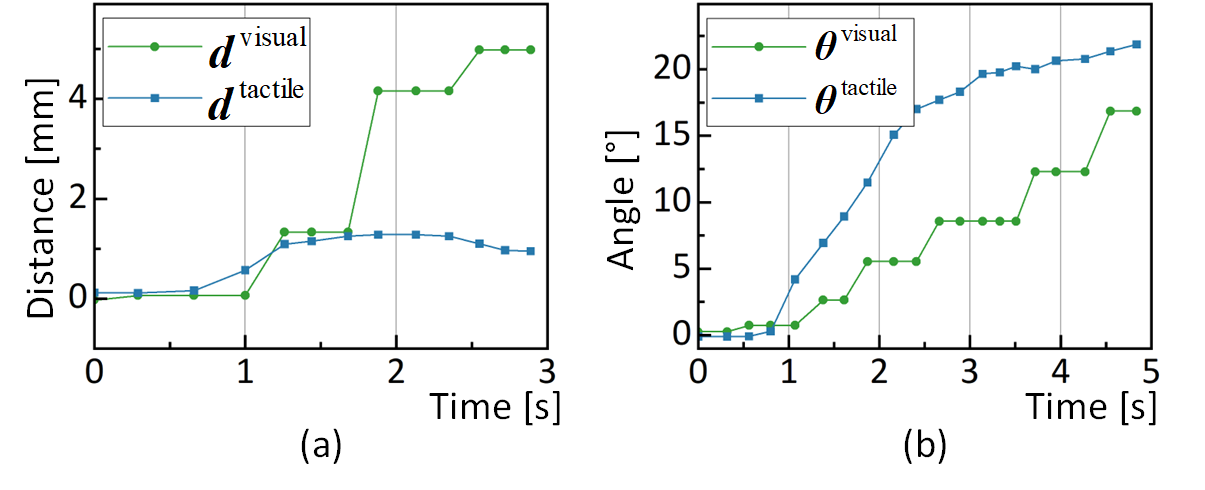}
    \caption{Measurement results from vision and tactile sensing. (a) Estimated object slip distance. (b) Estimated object rotation angle.}
    \label{fig::sensing results}
\end{figure}

To characterize the responses of visual and tactile perception during in-hand motion, two externally induced maneuvers were performed. The nominal 10 mm and $10^\circ$ motions specify the intended manipulation magnitudes rather than independently measured ground truth. The resulting traces are therefore used to examine the response characteristics of the two sensing modalities, rather than to quantify their absolute measurement accuracy.
\begin{itemize}
    \item \textbf{Experiment 1: Sliding.} This experiment compared the visual and tactile measurements during in-hand sliding. An external force was applied to induce a nominal 10 mm translation of the object within the gripper, and the corresponding displacement measurements were recorded (Fig.~\ref{fig::sensing results}(a)).
    \item \textbf{Experiment 2: Rotation.} This experiment compared the visual and tactile measurements during in-hand rotation. An external force was applied to induce a nominal $10^\circ$ rotation of the object within the gripper, and the corresponding rotational measurements were recorded (Fig.~\ref{fig::sensing results}(b)).
\end{itemize}

The key observations are summarized as follows:
\begin{itemize}
    \item \textbf{Slip Estimation:} Tactile sensing exhibits superior sensitivity to slip onset compared to vision, reacting immediately to tangential disturbances. However, its accuracy degrades under large-magnitude slip due to sensor saturation and non-linear deformation (Fig. \ref{fig::sensing results}(a)).
    \item \textbf{Rotation Estimation:} Visual perception provides stable global rotation estimates. In contrast, tactile rotation estimation is highly sensitive to the assumed center of rotation; slight shifts in contact geometry can lead to significant angular errors (Fig. \ref{fig::sensing results}(b)).
\end{itemize}

The stepwise visual trajectories in Fig.~5 are consistent with the update mechanism of the archived ArUco processing pipeline, which accumulates six valid marker detections before publishing a new filtered pose estimate. Consequently, the processed visual output may remain unchanged across multiple logging instants, even when the underlying motion is continuous. This behavior should not be interpreted as the raw camera measuring only discrete physical movements. Since the original acquisition-script version and frame-level timestamps were not preserved, the contribution of this batching mechanism to the plotted plateaus cannot be isolated conclusively.
Based on these observations, a fusion strategy is adopted that relies on vision for rotation estimation while adaptively combining visual and tactile data for translation (slip) estimation. The fused state is defined as:
\begin{align}
    \label{equ::sensor fussion of rotation}
    \hat{o}_{i|\boldsymbol{s}_i} &= \big( \theta(\boldsymbol{s}_i), \boldsymbol{d}(\boldsymbol{s}_i) \big), \\
    \theta(\boldsymbol{s}_i) &= \theta^\mathrm{visual}(\boldsymbol{s}_i),\\
    \boldsymbol{d}(\boldsymbol{s}_i) &= \boldsymbol{d}^\mathrm{visual}(\boldsymbol{s}_i) + \gamma^\mathrm{slip} \gamma^\mathrm{count} \boldsymbol{d}^\mathrm{tactile}(\boldsymbol{s}_i).
    \label{equ::sensor fussion}
\end{align}
Here, the tactile contribution is modulated by two dynamic weights to mitigate the errors identified above:
\begin{equation} 
    \label{equ::weight of slip}
    \gamma^\mathrm{slip} = \gamma^{\mathrm{low}} + (\gamma^\mathrm{high} - \gamma^{\mathrm{low}}) \min \left(\frac{\|\boldsymbol{d}^\mathrm{tactile}\|}{H_0}, 1.0 \right),
\end{equation}
\begin{equation}
    \label{equ::weight of continuity}
    \gamma^\mathrm{count} = 0.5^{c-1}.
\end{equation}
The slip-dependent weight $\gamma^\mathrm{slip}$ increases tactile influence when significant contact displacement is detected ($\|\boldsymbol{d}^\mathrm{tactile}\|$), ensuring prompt reaction to slippage. Conversely, the temporal decay weight $\gamma^\mathrm{count}$ is empirically designed as an exponential decay function to reduce tactile reliance over consecutive identical actions ($c$), preventing error accumulation from the no-slip approximation during continuous motion.

\subsection{Evaluation of Gripper Action}
Directly optimizing the similarity metric in Equation (\ref{equ::action evaluation based on similarity}) is insufficient, as it tends to bias the system toward large-amplitude actions. Since larger actions naturally induce greater state changes, they may yield high similarity scores even if the motion direction is suboptimal. This can lead to a greedy strategy that prioritizes magnitude over directional efficiency.

To address this, the evaluation metric is reformulated to assess the efficiency of the action—defined as the task progress per unit of actuation. The refined metric is given by:

\begin{equation}
    \label{equ::action evaluation with magnitude}
    G(\boldsymbol{s}_i,\boldsymbol{a}_i) = \big( \mathrm{Sim}(\hat{o}_{i+1|\boldsymbol{s}_i,\boldsymbol{a}_i}, \mathbb{E} [o]) \big)/{\hat{l}}.
\end{equation}
where $\hat{l}$ represents the normalized action amplitude. This normalization is essential for handling the heterogeneous action space (e.g., translation in mm vs. rotation in degrees). Specifically, $\hat{l}$ is computed as:
\begin{equation}
    \label{equ::action normalized}
    \hat{l} = l_i / l^\mathrm{range}_i,
\end{equation}
where $l_i$ is the raw amplitude of the $i$-th component and $l^\mathrm{range}_i$ is its physical limit. This formulation ensures that actions are evaluated based on their directional alignment with the task goal rather than their magnitude, promoting a balanced selection across different degrees of freedom.

\hypertarget{sec::action adjustment}{}

\section{Action Adjustment}

The utility of a manipulation action is inherently dynamic, depending heavily on the evolving object pose and external contact conditions. While the previous section established a metric for evaluating the immediate outcome of an action, relying solely on post-execution evaluation is insufficient for real-time control. Direct search within the action space based on immediate feedback leads to excessive exploration and fails to account for the long-term impact of actions.

To address this, this section proposes an online action adjustment strategy based on \textit{local reward estimation}. The action utility formulation is extended in three key aspects to enhance prediction accuracy and efficiency:
(1) \textbf{Local Linearization}: By analyzing state differentials, the utility of local actions is estimated without exhaustive sampling;
(2) \textbf{Uncertainty Utility}: An exploration term is introduced to mitigate the unreliability of predictions caused by temporal decay;
(3) \textbf{Future Utility}: A long-term perspective is incorporated, crediting actions that may have low immediate returns but facilitate subsequent task progress.

Finally, a stepwise reward update mechanism is established. This closed-loop process continuously refines utility estimates by integrating immediate feedback with historical predictions, enabling stable and efficient action selection under uncertainty.

\subsection{Overview of Action Utility Modeling}
\begin{figure}[t]
    \centering
    \includegraphics[width = 1.0\linewidth]{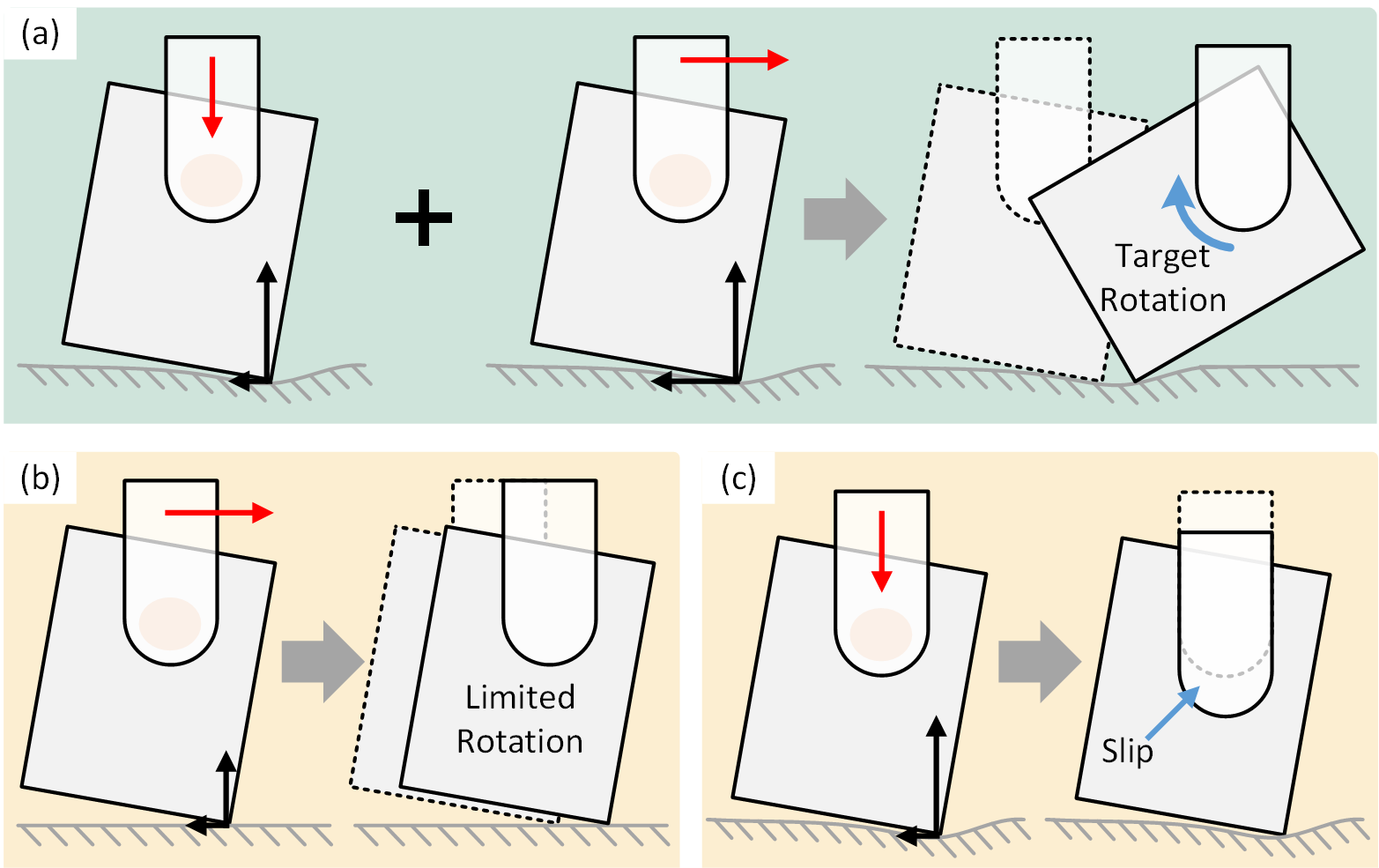}
    \caption{Illustration of the analysis of future action utility. (a) Effects of sequential execution of two types of actions. (b) Effect of executing the translational action alone. (c) Effect of executing the downward motion alone.}
    \label{fig::combined effect analysis}
\end{figure}

Relying solely on immediate utility is insufficient for effective reorientation, as the optimal policy often requires sequential actions where early steps facilitate later success. 
Consider the scenario in Fig. \ref{fig::combined effect analysis}: a direct lateral push may cause excessive slip. However, an initial downward motion increases normal force and friction, enabling a subsequent lateral push to effectively trigger rotation. This demonstrates that an action's value is not solely determined by its immediate kinematic result, but also by its \textit{coupling} with future actions.

Consequently, a robust action evaluation framework must account for three distinct aspects:
\begin{enumerate}
    \item \textbf{Immediate Utility:} The direct contribution of the action to the task progress.
    \item \textbf{Future Utility:} The potential of the action to create favorable conditions for subsequent steps, capturing the inter-action coupling and long-term impact.
    \item \textbf{Uncertainty utility:} The incentive to explore actions with high uncertainty to prevent getting stuck in local optima due to temporal decay of information.
\end{enumerate}

Mathematically, the total reward function is formulated as a weighted sum of these components:
\begin{equation}
    \label{equ::reward function}
    R(s_i, \boldsymbol{a})=U_\mathrm{n}(s_i, \boldsymbol{a}) + \beta_1 U_\mathrm{f}(s_i, \boldsymbol{a}) + \beta_2 U_\mathrm{e}(s_i, \boldsymbol{a})
\end{equation}
where $U_\mathrm{n}$, $U_\mathrm{f}$, and $U_\mathrm{e}$ represent the immediate, future, and uncertainty utilities, respectively, and $\beta_1, \beta_2$ are their weighting coefficients.

\hypertarget{sec::local linearization}{}
\subsection{Local Linearization of Action Evaluation Metrics}

Evaluating the metric $G(\boldsymbol{a})$ over the entire high-dimensional action space is computationally prohibitive for real-time control. To address this challenge, a local linearization strategy is introduced that leverages the continuity and smoothness of $G(\boldsymbol{a})$ in the vicinity of a given action. This approach simplifies the action space locally and significantly reduces the number of samples required for estimating $G(\boldsymbol{a})$.

First, the evaluation metric is reformulated to focus on state changes rather than absolute states. This differential form aligns with the continuous nature of manipulation physics:
\begin{equation}
    \label{equ::action evaluation expanded}
    G(s_i,\boldsymbol{a}_i) = \mathrm{Sim}(\Delta \hat{o}_{i|s_i,\boldsymbol{a}_i}, \mathbb{E} [\Delta o_i]) / \hat{l},
\end{equation}
where $\Delta \hat{o}$ is the predicted observation change and $\mathbb{E} [\Delta o_i]$ is the expected change derived from task sub-goals. Specifying the similarity measure as a dot product, the metric decomposes into rotation and slip components:
\begin{equation}
\label{equ::action evaluation}
G(s_i,\boldsymbol{a}_i)
=
\frac{
    w_{\theta}\Delta\theta\,e_0(s_i)
    +
    w_d\Delta\boldsymbol{d}
    \cdot\boldsymbol{u}_0(s_i)
}{
    \hat{l}
}.
\end{equation}
where $w_{\theta}$ and $w_d$ are scaling coefficients for the rotational and translational responses, respectively. They account for the different numerical scales and physical units of the two motion components when constructing the scalar action evaluation metric.
Here, $e_0$ and $\boldsymbol{u}_0$ are the instantaneous sub-targets for rotation alignment and slip minimization, updated in real-time as:
\begin{align}
    \label{equ::sub goals (rotation)}
    e_0 (s_i) &= \frac{\theta_\mathrm{target} -\theta(s_{i})}{| \theta_\mathrm{target} -\theta(s_{i}) |}, \quad
    \boldsymbol{u}_0 (s_i) = - \frac{\boldsymbol{d} (s_i)}{\|\boldsymbol{d} (s_i)\|}.
\end{align}

Both sub-targets $e_0(s_i)$ and $\boldsymbol{u}_0(s_i)$ are updated in real time. This update mechanism offers several advantages. First, it allows the system to re-evaluate and adjust its actions at every step based on current feedback, rather than relying on a fixed long-term plan, thereby reducing the risk of error accumulation over time. Second, it enables the system to react promptly to unexpected disturbances in unstructured environments.

Any action $\boldsymbol{a}$ is decomposed into an amplitude $\hat{l}$ and a direction $\boldsymbol{e}_a$. When the action amplitude $\hat{l}$ tends to zero, both $\Delta \theta$ and $\Delta \boldsymbol{d}$ approach zero, as expressed by
\begin{equation} 
    \label{equ::continuity} 
    \lim_{\hat{l} \rightarrow 0} (\Delta \theta, \Delta \boldsymbol{d}) = (0,\boldsymbol{0}). 
\end{equation}

Although continuity does not imply linearity, quasistatic contact interactions exhibit smooth (differentiable) mappings from action amplitude to object motion, as contact geometry and frictional responses vary smoothly under small perturbations. Under this differentiability assumption, a first-order Taylor expansion with respect to $\hat{l}$ yields the following linear approximations:
\begin{align}
    \label{equ::sensor Jacobian-1}
    \Delta \theta(\boldsymbol{a}) \approx J_\theta \hat{l}, \quad
    \Delta \boldsymbol{d}(\boldsymbol{a}) \approx \boldsymbol{J}_d \hat{l},
\end{align}
where $J_\theta$ and $\boldsymbol{J}_d$ are the local Jacobians representing the sensitivity of rotation and slip to the action direction. 

It is important to note that this linearity assumption may be violated in scenarios involving abrupt contact mode transitions (e.g., stick-slip) or geometric discontinuities, leading to estimation errors in the predicted utility. However, as detailed in Section \hyperlink{sec::utility update and action selection}{\uppercase\expandafter{\romannumeral5}-D}, the proposed framework compensates for these inaccuracies through a closed-loop correction mechanism: the actual outcome of an executed action is used to update the utility model immediately. Furthermore, the uncertainty utility term $U_\mathrm{e}$ encourages the exploration of actions with low confidence, preventing the system from being misled by erroneous local estimates in non-smooth regions.

Substituting these into Equation (\ref{equ::action evaluation}) yields the linearized metric:
\begin{equation}
    \label{equ::action evaluation with Jocobian}
    G(\boldsymbol{a}) |_{\hat{l} \rightarrow 0} = J_\theta e_0 + \boldsymbol{J}_d \cdot \boldsymbol{u}_0.
\end{equation}

This linearization offers two critical advantages for efficiency:
1) \textbf{Amplitude Independence:} The metric depends solely on the action direction, reducing the search space from a volume to a unit sphere.
2) \textbf{Symmetric Inference:} The linearity implies $G({-\boldsymbol{a}}) \approx - G(\boldsymbol{a})$, allowing the utility of a reverse action to be inferred directly without additional sampling.

\subsection{Temporal Action Utility Modeling: Uncertainty and Future Utility}
\begin{figure*}[t]
    \centering
    \includegraphics[width = 0.9\linewidth]{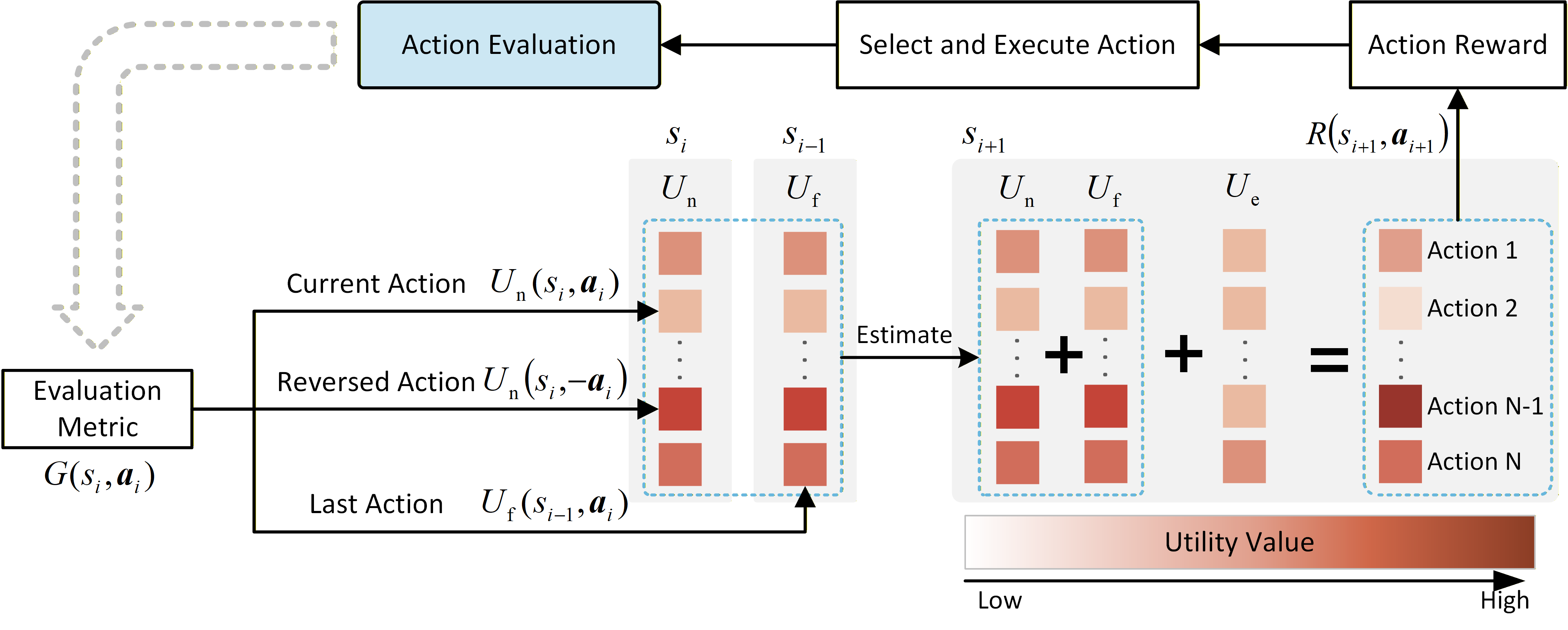}
    \caption{Action reward update pipeline. The evaluation metric obtained after action execution is used to update the immediate utility, which in turn updates the utilities of the current, reverse, and previous actions. These updated utilities are used to estimate the expected rewards for candidate actions at the next step $s_{i+1}$, guiding the selection and execution of the next action in an iterative loop.}
    \label{fig::value update}
\end{figure*}

Action utility is inherently time-dependent. This introduces two critical challenges for online estimation: (1) \textit{Information Decay}, where the confidence in the estimated utility of an action decreases if it remains unexecuted for a long period; and (2) \textit{Long-term Dependency}, where the value of an action lies not in its immediate result but in its ability to facilitate future success. To address these, the reward function is augmented with two temporal components.

To prevent the system from converging prematurely to local optima due to outdated information, an uncertainty utility $U_\mathrm{e}$ is introduced. Inspired by the Upper Confidence Bound (UCB) algorithm, this term encourages the re-exploration of actions that have not been visited recently:
\begin{equation}
    \label{equ::utility of exploration}
    U_\mathrm{e}(s_i, \boldsymbol{a}) = \sqrt{\frac{\ln {t_i}}{N_{\boldsymbol{a}} + 1}},
\end{equation}
where $t_i$ denotes the current time step, and $N_{\boldsymbol{a}}$ is the number of times action $\boldsymbol{a}$ has been executed. This term decays as an action is frequently selected, naturally balancing exploration and exploitation.

To capture the coupling effect where a current action enables a high-reward subsequent action, the future utility $U_\mathrm{f}$ is introduced. This component propagates value backward in time. Specifically, the future utility of an action at state $s_i$ is updated based on the immediate utility observed at the subsequent state $s_{i+1}$:
\begin{align}
    \label{equ::utility of future value}
    U_\mathrm{f}(s_i, \boldsymbol{a}) \leftarrow 
    \begin{cases}
        (1-\alpha) U_\mathrm{f}(s_i, \boldsymbol{a}) \\
        \quad + \alpha G(s_{i+1}, \boldsymbol{a})
        & \text{if } U_\mathrm{n}(s_{i+1}, \boldsymbol{a}) \\
        & > \widetilde{U}_\mathrm{n}(s_{i+1}, \boldsymbol{a}), \\
        U_\mathrm{f}(s_i, \boldsymbol{a}) & \text{else}.
    \end{cases}
\end{align}
Here, the update is conditional: credit is assigned only if the subsequent immediate utility $U_\mathrm{n}(s_{i+1})$ exceeds a predicted baseline threshold $\widetilde{U}_\mathrm{n}$. This filtering mechanism prevents the propagation of noise or rewards from trivial state transitions.

The threshold $\widetilde{U}_\mathrm{n}$ is estimated via a linear regression of the utility trend using the latest $m+1$ recorded utility values (empirically, $m=2$), predicting the expected utility at $t_{i+1}$:
\begin{equation}
    \label{equ::utility of now (predict)}
    \widetilde{U}_\mathrm{n}(s_{i+1},\boldsymbol{a})
    =
    \bar{U}+\theta_1(t_{i+1}-\bar{t}).
\end{equation}
where $\theta_1$ is the slope determined by least squares fitting:
\begin{align} 
    \label{equ::utility of now (predict fit)}
    &\theta_1=\frac{\sum_{j=k-m}^k(t_{j}-\bar{t})\cdot({U}_{\mathrm{n}}(s_j, \boldsymbol{a})-\bar{U})}{\sum_{j=k-m}^k(t_{j}-\bar{t})^2}, \\
    &\bar{t}=\frac{1}{m+1}\sum_{j=k-m}^k t_{j}, \, \bar{U}=\frac{1}{m+1}\sum_{j=k-m}^k {U}_\mathrm{n}(s_j, \boldsymbol{a}). \notag
\end{align} 

This trend-based thresholding ensures that the system distinguishes between genuine task progress and random fluctuations, thereby robustly identifying actions that contribute to long-term success.

\hypertarget{sec::utility update and action selection}{}
\subsection{Utility Update and Action Selection}


Accurate utility assessment requires physical interaction; the true value of an action is only revealed after execution. To bridge the gap between estimation and reality, a closed-loop stepwise reward update strategy is proposed, as illustrated in Fig. \ref{fig::value update}. This process iterates through prediction, execution, and correction.

(1) Utility prediction before execution. Before execution, the total expected reward $\hat{R}$ for each candidate action is synthesized from historical data and exploration incentives:
\begin{align}
    \label{equ::reward function with estimation}
    \hat{R}(s_i, \boldsymbol{a}) & = \hat{U}_\mathrm{n}(s_i, \boldsymbol{a}) + \beta_1 \hat{U}_\mathrm{f}(s_i, \boldsymbol{a}) + \beta_2  U_\mathrm{e}(s_i, \boldsymbol{a}),
\end{align}
where $\hat{U}_\mathrm{n}$ and $\hat{U}_\mathrm{f}$ are the most recent recorded immediate and future utilities, respectively, and $U_\mathrm{e}$ is the analytically computed exploration term (Equation (\ref{equ::utility of exploration})).

(2) Action execution. Based on the estimated total utility $\hat{R}(s_i, \boldsymbol{a})$ for all candidate actions, the action with the highest estimated utility is selected for execution. To balance efficiency with optimality, a "satisficing" strategy is adopted rather than searching for a global optimum at every step. The system prioritizes actions that are deemed "good enough" to progress the task. Let $M(s_i,\boldsymbol{a})
=\widehat{U}_\mathrm{n}(s_i,\boldsymbol{a})
+\beta_1\widehat{U}_\mathrm{f}(s_i,\boldsymbol{a})$
denote the predicted utility excluding
the exploration bonus. The selection logic is defined as:
\begin{equation}
\label{equ::utility-based action trial}
\boldsymbol{a}_i =
\begin{cases}
\displaystyle
\arg\max_{\boldsymbol{a}\in\mathcal{A}}
M(s_i,\boldsymbol{a}),
&
\displaystyle
\max_{\boldsymbol{a}\in\mathcal{A}}
M(s_i,\boldsymbol{a})>\mu_0,
\\[2mm]
\displaystyle
\arg\max_{\boldsymbol{a}\in\mathcal{A}}
\widehat{R}(s_i,\boldsymbol{a}),
&
\mathrm{otherwise}.
\end{cases}
\end{equation}
\begin{equation}
    {R}_\mathrm{0}(s_i, \boldsymbol{a}) = \mu_0 + \beta_2 U_\mathrm{e}(s_i, \boldsymbol{a}).
\end{equation}
Here, ${R}_\mathrm{0}(s_i, \boldsymbol{a}) = \mu_0 + \beta_2 U_\mathrm{e}(s_i, \boldsymbol{a})$ serves as a dynamic satisfaction threshold. If an action's predicted utility exceeds this threshold, it is executed immediately to save computation. Otherwise, the action with the highest predicted utility is selected.

(3) Utility update after execution. Upon execution, the actual outcome is measured to refine the utility model. Three specific updates are performed.

\textbf{Immediate Utility Update:} The immediate utility $U_\mathrm{n}$ is updated using an exponential moving average to smooth out sensor noise:
\begin{equation}
    \label{equ::utility of now}
    U_\mathrm{n}(s_i, \boldsymbol{a}) \leftarrow (1-\alpha) U_\mathrm{n}(s_k, \boldsymbol{a}) + \alpha G(s_i, \boldsymbol{a}),
\end{equation}
where $G(s_i, \boldsymbol{a})$ is the measured evaluation metric, and $s_k$ denotes the state at the previous execution of action $\boldsymbol{a}$.

\textbf{Symmetric Correction:} Leveraging the local linearity established in Section (\hyperlink{sec::local linearization}{\uppercase\expandafter{\romannumeral5}-B}), the utility of the reverse action $-\boldsymbol{a}$ is simultaneously updated without additional sampling:
\begin{align}
    \label{equ::utility update: reverse action utility}
    U_\mathrm{n}(s_i, {-\boldsymbol{a}}) \leftarrow (1-\alpha) U_\mathrm{n}(s_i, {-\boldsymbol{a}}) + \alpha G(s_i, {-\boldsymbol{a}}),
\end{align}
where $G(s_i, {-\boldsymbol{a}})$ is inferred from the symmetric property of the linearized metric.

\textbf{Future Utility Backpropagation:} The future utility $U_\mathrm{f}$ of the \textit{previous} action is updated based on the current step's success, as defined in Equation (\ref{equ::utility of future value}).

The complete iterative process of prediction, selection, and update is summarized in Algorithm \ref{alg::action optimization}.

\section{Experiment}

\begin{algorithm}[t]
    \caption{Object Reorientation Strategy}
    \label{alg::action optimization}
    \KwData{Target pose $\theta_{\mathrm{goal}}$}
    \KwResult{Optimized action sequence}
    
    Initialize $\theta_{\mathrm{current}}$, action set $\mathcal{A}$, and utility values\;

    \While{$\|\theta_{\mathrm{current}} - \theta_{\mathrm{goal}}\| > \epsilon$}{
        \tcc{Step 1: Utility Prediction}
        \For{$a \in \mathcal{A}$}{
            $R(a) \gets$ ComputeReward$(a)$\;  
        }

        $\boldsymbol{a}_{\mathrm{selected}} \gets$ SelectAction$(\mathcal{A}, R)$\;

        \tcc{Step 2: Action Execution}
        Execute$(\boldsymbol{a}_{\mathrm{selected}})$ for time step $\Delta t$\;
        $\Delta\theta, \Delta\boldsymbol{d}  \gets$ MeasureObjectMovement$()$\;
        $G(\boldsymbol{a}_{\mathrm{selected}})  \gets$ UpdateEvaluation$(\boldsymbol{a}_{\mathrm{selected}}, \Delta\theta, \Delta\boldsymbol{d})$\;

        \tcc{Step 3: Utility Update}
        $U_\mathrm{n}(\boldsymbol{a}_{\mathrm{selected}}) \gets$ UpdateImmeUtility$(\boldsymbol{a}_{\mathrm{selected}})$\;
        $U_\mathrm{n}({-\boldsymbol{a}}_{\mathrm{selected}}) \gets$ UpdateImmeUtility$({-\boldsymbol{a}}_{\mathrm{selected}})$\;
        $U_\mathrm{f}(\boldsymbol{a}_{\mathrm{prev}}) \gets$ UpdateFutureUtility$(\boldsymbol{a}_{\mathrm{prev}})$\;
        
        \For{$a \in \mathcal{A}$}{
            $U_\mathrm{e}(\boldsymbol{a}) \gets$ UpdateExplorationUtility$(a)$\;
        }

        $\theta_{\mathrm{current}} \gets$ GetCurrentPose()\;
    }
\end{algorithm}

This section evaluates the performance and robustness of the proposed framework through a series of experiments. The evaluation focuses on three key aspects: 
(1) \textbf{Adaptability Assessment}: verifying the method's generalization capabilities across diverse objects, materials, and environmental contact conditions; 
(2) \textbf{Ablation Study}: analyzing the individual contributions of perception modalities, utility gradients, and reward components to the overall system performance; 
(3) \textbf{Real-world Demonstration}: validating practical applicability through a challenging case study involving the reorientation of a cabled charger under complex external disturbances.

\subsection{Experiment Setup}
The experimental system integrates visual and tactile feedback for state estimation. Object pose is tracked using a camera system combined with ArUco markers, achieving a position estimation error within 0.1 mm and an orientation error within 0.02$^\circ$ at a frequency of 5 Hz. Tactile feedback is provided by the Tac3D sensor \cite{zhang2022tac3d}, which features a surface embedded with a $20\times20$ marker array. The sensor provides dense deformation measurements with a marker tracking accuracy better than 0.03 mm at 30 Hz.

\subsection{Adaptability Evaluation}

This experiment evaluates the generalization capability of the proposed method across variations in object shape and environmental contact conditions. The primary metric is the task success rate. A trial is deemed successful if: (1) the final orientation error is less than 5°, and (2) the finger contact displacement remains within 30 mm (ensuring the object remains within the grasp).

\begin{figure}[t]
    \centering
    \includegraphics[width = 1\linewidth]{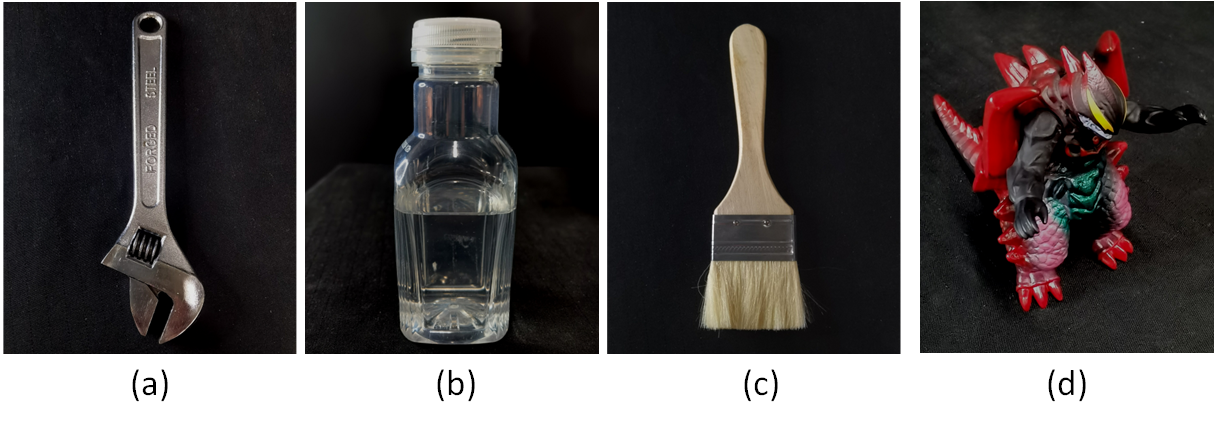}
    \caption{Representative evaluation objects with different contact characteristics: wrench (rigid point contact), half-filled bottles (rigid line contact and variable center of gravity), brushes (nonlinear multi-point environmental contact), and toys (multi-point contact).}
    \label{fig::baseline scenarios}
\end{figure}

As illustrated in Fig.~\ref{fig::baseline scenarios}, four representative objects were selected to cover distinct contact characteristics: a wrench (rigid point contact), a half-filled bottle (variable center of gravity), a brush (nonlinear compliant contact), and a toy (multi-point contact). Additionally, a low-friction scenario was introduced to test robustness against environmental uncertainty. In total, 5 experimental scenarios were conducted, with 10 trials per scenario. In each trial, the object's initial grasp pose was randomized. Key metrics recorded included success rate and the number of adjustment steps taken.

To ensure fair comparison across trials with varying initial poses, the number of action steps is normalized. Let $\Delta \mathbf{P}$ denote the angular magnitude of the required reorientation. The normalized step count $T_{\mathrm{norm}}$ is scaled to a reference rotation of $90^\circ$:
\begin{equation}
    T_{\mathrm{norm}}=\frac{T_{\mathrm{raw}}}{\|\Delta\mathbf{P}\|} \times 90^\circ,
\end{equation}
where $T_{\mathrm{raw}}$ is the actual step count.

The proposed method is compared against a Proximal Policy Optimization (PPO) baseline trained in a MuJoCo-based simulation using seven objects under varying friction coefficients (0.5, 1.0, 1.5). The PPO policy controls planar translations along the $y$ and $z$ axes, whereas the proposed controller also permits grasp-force adjustment. Therefore, PPO serves as a simulation-trained reference baseline rather than a strictly action-space-matched comparison. Further implementation details are provided in Appendix~\ref{app:ppo}.



\begin{table}[t]
    \centering
    \caption{Reorientation Performance Comparison}
    \label{tab::baseline compare}
    \begin{tabular}{cccc}
    \toprule
    \textbf{Scenario} & \textbf{Method} & \textbf{Success/Trials}~$\uparrow$  & $\boldsymbol{T}_{\text{norm}}$~$\downarrow$ \\
    \midrule
    \multirow{2}{*}{Wrench} & Ours & 10/10  & \textbf{60.99} \\
    \multirow{2}{*}{}       & PPO  & 10/10  & 81.58 \\
    \cmidrule{0-1}
    \multirow{2}{*}{Half-filled bottle} & Ours & 10/10 & 61.97 \\
    \multirow{2}{*}{}                   & PPO  & 10/10 & \textbf{54.13} \\
    \cmidrule{0-1}
    \multirow{2}{*}{Brush}  & Ours & \textbf{10/10} & 59.43 \\
    \multirow{2}{*}{}       & PPO  & 9/10  & \textbf{50.95} \\
    \cmidrule{0-1}
    \multirow{2}{*}{Toy}    & Ours & \textbf{10/10} & 50.12 \\
    \multirow{2}{*}{}       & PPO  & 8/10           & \textbf{43.46} \\
    \cmidrule{0-1}
    \multirow{2}{*}{Wrench (low friction) } & Ours & \textbf{10/10} & \textbf{68.25} \\
    \multirow{2}{*}{}                       & PPO  & 9/10  & 75.17 \\
    \bottomrule
    \end{tabular}
    \vspace{3pt}
\end{table}

Quantitative results are summarized in Table \ref{tab::baseline compare}. The proposed method achieved a 100\% success rate (10/10) across all scenarios, consistently outperforming the PPO baseline.

\begin{figure}[t]
    \centering
    \includegraphics[width = 1.0\linewidth]{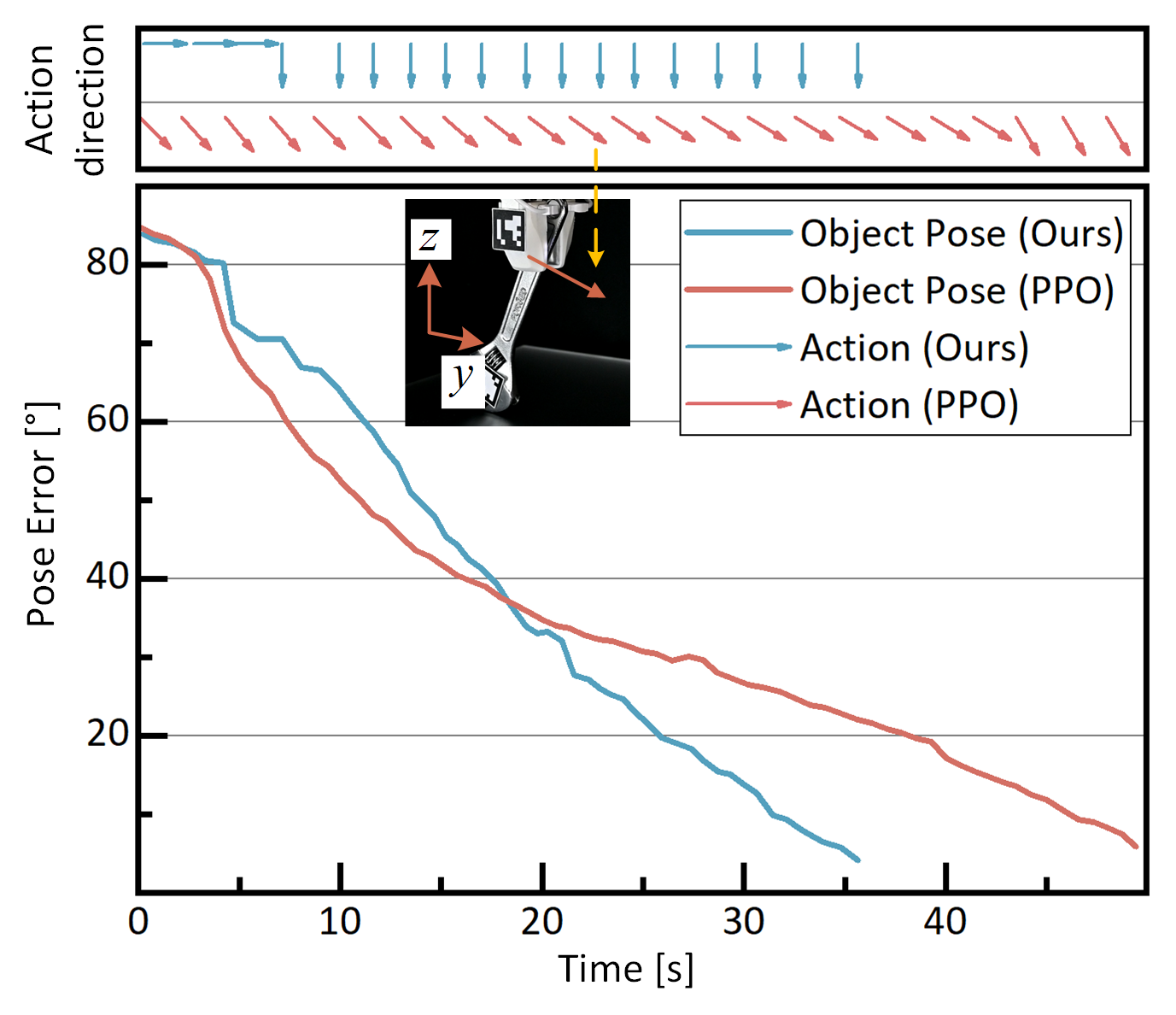}
    \caption{Comparison of methods in reorienting the wrench. In the top figure, arrows indicate the actions of the robotic hand at different time steps. The bottom figure shows the evolution of orientation error over time.}
    \label{fig::baseline compare}
\end{figure}

To understand the performance gap, the behavioral differences are analyzed using the wrench task (Fig. \ref{fig::baseline compare}).
\begin{itemize}
    \item \textbf{PPO (Conservative Strategy):} The PPO policy adopts a conservative approach, incrementally increasing horizontal motion while suppressing vertical action to minimize relative slip. While this enhances robustness against friction uncertainty, it often results in inefficient, prolonged adjustments.
    \item \textbf{Ours (Adaptive Strategy):} In contrast, the proposed method dynamically adjusts its strategy online. Upon detecting that vertical action effectively drives reorientation (especially in low-friction conditions), it rapidly switches dominance to that direction. This ability to exploit slip, rather than suppress it, significantly improves reorientation efficiency.
\end{itemize}

In summary, the proposed method demonstrates robust zero-shot adaptability. Unlike the PPO baseline, which requires extensive pre-training and retraining when robot kinematics change, the proposed approach relies on real-time perception and local evaluation. This allows it to handle novel objects and disturbances effectively without prior knowledge.

\subsection{Ablation and Component Analysis}

This section analyzes the individual contributions of three core components: (1) visual-tactile fusion, (2) utility gradients, and (3) the specific terms within the reward function. To isolate these effects, ablation studies were conducted using the wrench reorientation task with 10 randomly generated initial poses. The robot's action space was restricted to planar translation and grasping force adjustment. This constrained action subspace preserves the essential challenges of the multi-DoF manipulation problem while making the resulting behaviors easier to visualize and analyze.

\setcounter{topnumber}{1}
\begin{table}[t]
    \centering
    \caption{Ablation Study Results}
    \label{tab::ablation_results}
    \footnotesize
    \setlength{\tabcolsep}{3pt}
    \begin{tabular}{lccc}
    \toprule
    \textbf{Method} & \shortstack{\textbf{Success/}\\\textbf{Trials}~$\uparrow$} & $\boldsymbol{T_{\mathrm{norm}}}$~$\downarrow$ & \shortstack{\textbf{Early} $\boldsymbol{\Delta S_{20}}$\\(mm)~$\downarrow$} \\
    \midrule
    Full Method               & \textbf{10/10} & \textbf{77.55} & \textbf{$5.25\pm2.91$} \\
    w/o Tactile Reward         & 9/10          & 79.01          & $10.10\pm7.81$ \\
    w/o Gradient               & 3/10          & --             & -- \\
    w/o Future Utility         & 7/10          & 87.38          & -- \\
    w/o uncertainty utility    & 7/10          & 83.83          & -- \\
    Random Greedy              & 3/10          & --             & -- \\
    \bottomrule
    \end{tabular}
    \vspace{3pt}
    \begin{tablenotes}
    \footnotesize
    \item \textbf{Note:}
    $T_{\mathrm{norm}}$ averages successful trials only.
    $S$ denotes a vision-based estimate of in-hand slip,
    computed from deviations from ideal pure rotation.
    Early $\Delta S_{20}=S_{20}-S_0$ measures the
    estimated slip increase over the first 20 effective
    actions (mean $\pm$ SD; $n=10$ per method,
    including failures).
    Dashes indicate unavailable results.
    \end{tablenotes}
    \vspace{-3mm}
\end{table}

\begin{figure}[!t]
    \centering
    \includegraphics[width=\columnwidth]
    {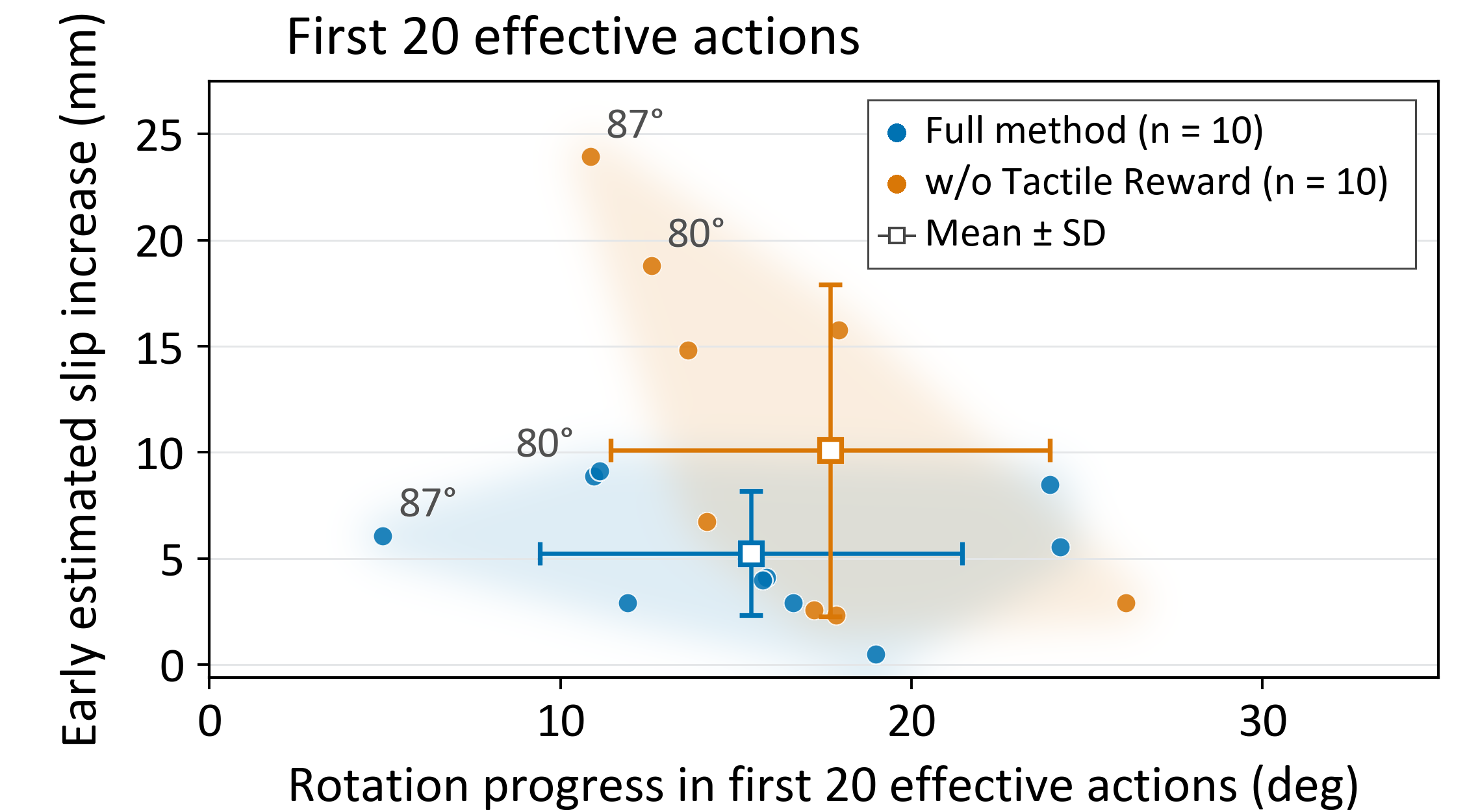}
    \caption{Multi-trial comparison of the full method
    and the ablation without tactile reward
    (10 trials each). Estimated slip increase
    versus rotation progress over the first
    20 effective actions. Each point represents
    one trial, while squares and error bars
    denote mean $\pm$ standard deviation.
    The hollow diamond marks the retained
    failed trial. Slip is estimated from vision
    as deviation from ideal pure rotation,
    rather than directly measured cumulative
    sliding at the finger--object interface.}
    \label{fig::tactile_reward_multitrial}
\end{figure}

\textbf{The impact of visual/tactile fusion.}

\begin{figure*}[t]
    \centering
    \includegraphics[width = 0.82\linewidth]{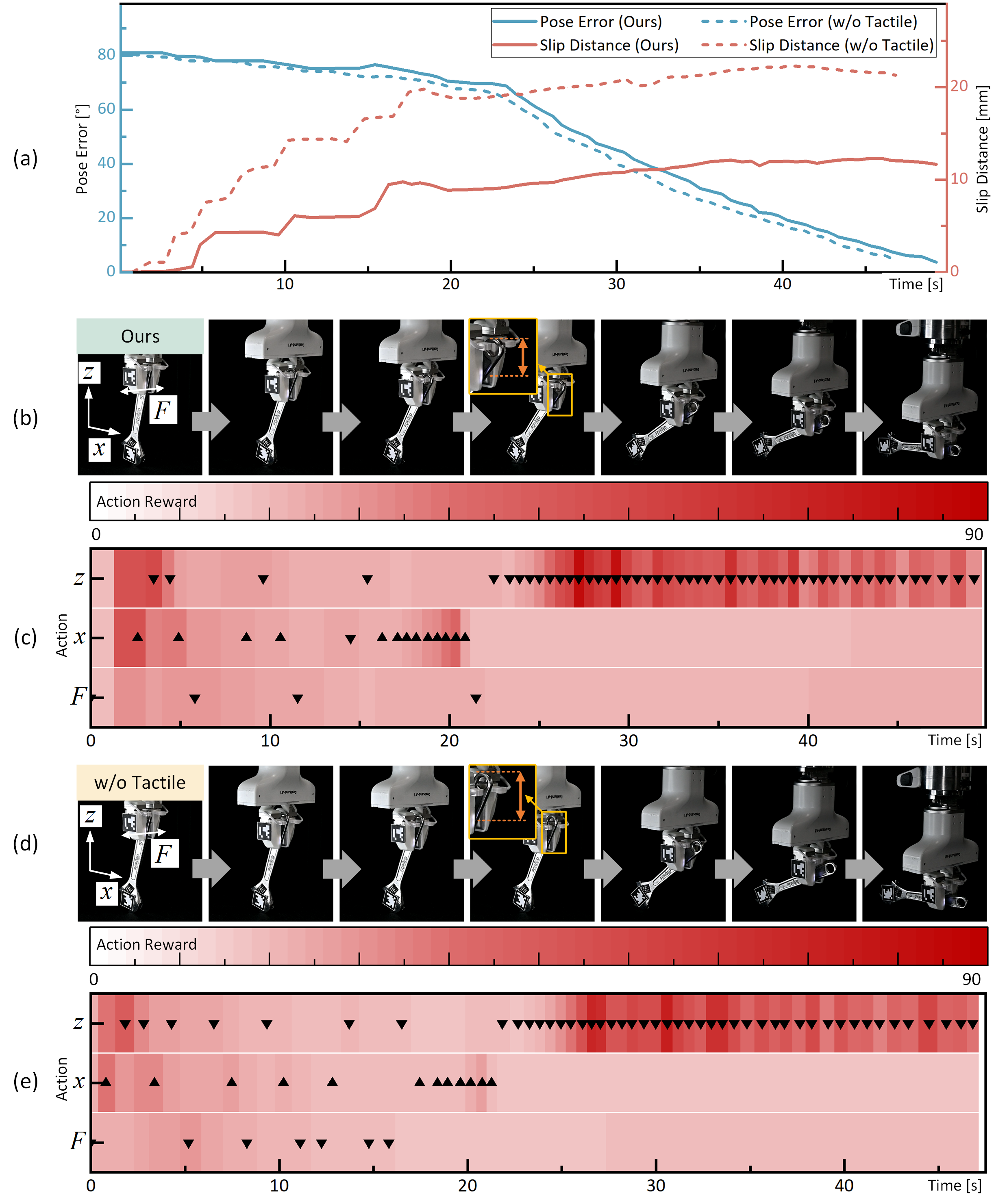}
    \caption{Comparison between the proposed method and the ablation without tactile contribution to action evaluation. (a) Evolution of orientation error and vision-estimated slip over time for both methods. (b) Snapshots at different time steps using the method. (c) Action rewards over time: the background color indicates the reward magnitude (darker denotes higher reward), and the triangles represent the selected actions, with their orientation indicating the action direction (upward = positive along the corresponding degree of freedom). (d), (e) Results of the ablation.}
    \label{fig::ablation feedback}
\end{figure*}

The proposed visual-tactile framework was compared against an ablation without tactile contribution to action evaluation. As shown in Table \ref{tab::ablation_results}, the ablation achieved a success rate of 9/10 with a normalized step count of 79.01, compared with 10/10 and 77.55 for the full method. The corresponding trajectory-level differences in orientation convergence and contact slip are further illustrated in Fig. \ref{fig::ablation feedback}.

To further examine the trajectory-level differences between the two methods, the evolution of orientation error, contact slip, and action selection is shown in Fig. \ref{fig::ablation feedback}. In this representative run, the ablation reaches the orientation threshold slightly earlier, whereas it accumulates substantially greater contact slip. The full method also exhibits fewer exploratory actions along the $-z$ direction and fewer grasp-force reductions during the early stage of reorientation.

Both types of actions--downward movement in $-z$ and frequent force reduction--are associated with increased finger slippage in this representative execution, as highlighted in the yellow box in Fig. \ref{fig::ablation feedback} (b) and (d). Consequently, the full method exhibits a substantially lower final slip distance than the ablation. Note that Fig. \ref{fig::ablation feedback} shows representative single-run trajectories in wall-clock time, whereas Table \ref{tab::ablation_results} reports statistics over 10 trials, with $T_{\mathrm{norm}}$ denoting the normalized action-step count rather than execution time.


To examine whether this behavior extends beyond
a representative run, Fig.~\ref{fig::tactile_reward_multitrial}
summarizes ten trials per method. Over the first
20 effective actions, the full method exhibits
a smaller increase in estimated slip
($5.25\pm2.91$~mm versus $10.10\pm7.81$~mm),
while achieving slightly less rotation progress
($15.43\pm6.01^\circ$ versus
$17.69\pm6.25^\circ$). At the first attainment of a 20\% reduction in initial orientation error, the mean estimated
slip is $5.74$~mm for the full method and $9.74$~mm for the ablation. These results indicate that tactile reward helps reduce estimated in-hand slip during reorientation, although the benefit varies across initial conditions.

\textbf{The impact of utility gradients.}
The efficacy of the gradient-based utility formulation (Equation (\ref{equ::action evaluation with magnitude})) was evaluated. A direct comparison with a gradient-free method in a continuous action space is infeasible due to the infinite exploration possibilities. Therefore, to ensure a fair baseline, a discretized strategy was implemented with six fixed-amplitude actions per degree of freedom, uniformly distributed within the interval $[-\mathrm{range}, \mathrm{range}]$, where $\mathrm{range}$ corresponds to the maximum action amplitude permitted by the gradient-based method along that degree of freedom.

Table \ref{tab::ablation_results} shows that the gradient-based method significantly outperforms the discretized baseline (Success: 10/10 vs. 3/10). Even with the search space artificially reduced for the baseline, the gradient-based approach demonstrated superior efficiency. This confirms that local linearization effectively guides the search direction, avoiding the inefficient trial-and-error characteristic of discrete exploration.

\textbf{The impact of the utility function.}

The roles of Future Utility ($U_\mathrm{f}$) and uncertainty utility ($U_\mathrm{e}$) were analyzed by ablating them individually. Additionally, the proposed exploration strategy was compared against a "Random Greedy" baseline (30\% random action probability) to validate the design of $U_\mathrm{e}$. The results in Table \ref{tab::ablation_results} indicate:

\begin{enumerate}
    \item Necessity of Future Utility: Removing $U_\mathrm{f}$ increased the number of steps (77.55 $\to$ 87.38) and reduced the success rate. This confirms that immediate rewards alone are myopic; $U_\mathrm{f}$ is essential for crediting actions that yield low immediate progress but enable future success.
    \item Necessity of uncertainty utility: Removing $U_\mathrm{e}$ caused the success rate to drop to 7/10. Without the incentive to revisit neglected actions, the system became prone to getting stuck in local optima.
    \item Superiority over Random Exploration: The Random Greedy strategy performed poorly (3/10 success). This highlights that structured exploration based on information decay ($U_\mathrm{e}$) is far more effective than stochastic noise, as it directs exploration toward actions with high uncertainty rather than randomly repeating known suboptimal actions.
\end{enumerate}

\subsection{Demonstration: Reorientation of a charger with cable}

\begin{figure*}[t]
    \vspace{-5mm}
    \centering
    \includegraphics[width = 0.93\linewidth]{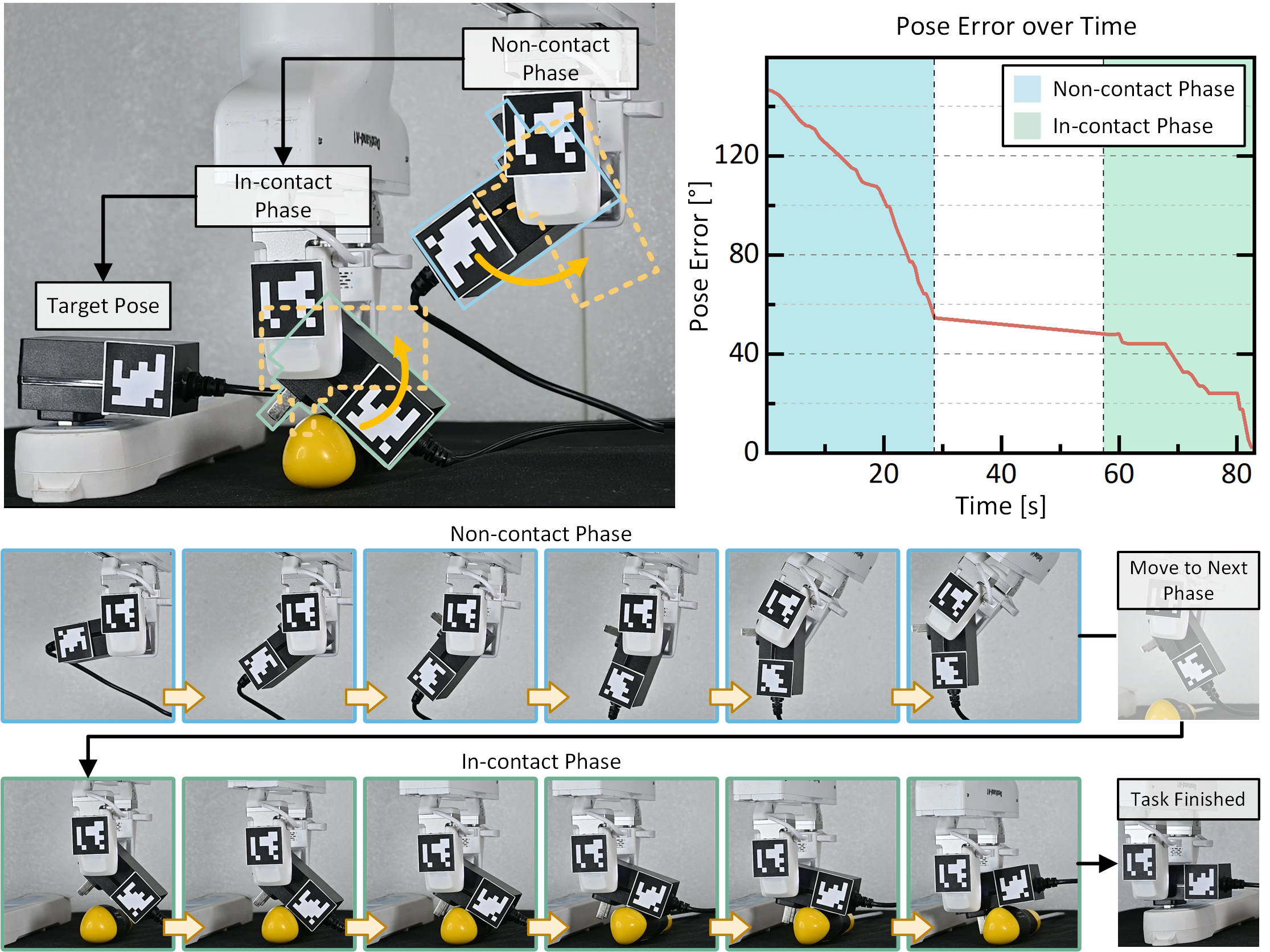}
    \caption{Task Progression and Pose Error Evolution During In-hand Reorientation of Cabled Charger.}
    \label{fig::demo charger}
    \vspace{-5mm}
\end{figure*}

To validate the practical applicability of the proposed method, a challenging case study was conducted: reorienting a charger with an attached cable. This task is distinct from rigid object manipulation due to the introduction of complex, time-varying external forces. During the mid-air phase, rotational actions are included in the candidate action set and selected online by the controller.

The objective is to adjust the charger's pose to align the plug for socket insertion. To facilitate this, rotational motion was included in the predefined candidate action set, allowing the online action-selection mechanism to choose rotational actions when they received favorable utility estimates under gravity and cable tension. The experiment introduces three specific challenges to test robustness:
\begin{itemize}
    \item \textbf{Nonlinear Cable Dynamics:} The dangling cable generates unpredictable, state-dependent tension and elastic rebound forces during motion.
    \item \textbf{Hybrid Interaction Modes:} The task is divided into two predefined phases. Phase I performs \textit{mid-air reorientation} using gravity and cable tension. Once the rotation angle reaches the predefined 50$^\circ$ threshold, the controller switches to Phase II for contact-aided reorientation using grounded support. Within each phase, actions are selected online from the corresponding candidate action set.
    \item \textbf{Unstructured Environmental Contact:} The ground support is provided by a loose, irregular obstacle rather than a fixed fixture. As the robot pushes against it, the obstacle itself may shift, introducing significant contact uncertainty.
\end{itemize}

As shown in Fig. \ref{fig::demo charger}, the proposed method successfully aligned the charger and completed the insertion. Within each predefined reorientation phase, the controller adapted its action selection online based on the observed action outcomes. In the representative mid-air execution, the rotational action was selected in 25 of 39 online action-selection steps.
In the mid-air phase, it effectively compensated for the nonlinear cable tension, suppressing the elastic rebound that typically destabilizes open-loop strategies. 
In the contact phase, it successfully leveraged the movable obstacle to complete the rotation without requiring a model of the obstacle's geometry or friction. 
This demonstration confirms that the proposed contact-agnostic framework can generalize to deformable objects and unstructured environments, effectively handling unmodeled dynamics through real-time perception and adjustment.

\subsection{Experimental Validation of Local Linearity}
\label{sec::local_linearity_validation}

To further examine the local linearity underlying
the action evaluation strategy, we conducted five
additional real-robot experiments involving
object--environment contact. A two-fingered gripper
held an object with a commanded grasp force of
0.8~N, while the robot translated against a fixed
external obstacle. The robot executed continuous
Cartesian motions at 8 or 16~mm/s. We recorded the actual end-effector displacement, ArUco-based visual observations, and tactile feedback. The reported orientation is the marker's image-plane projected angle.

The results reveal approximately linear
action--response relationships over local motion
intervals, despite the nonlinear nature of the
overall contact interaction. Three representative trajectories exhibited locally consistent projected-orientation response slopes of $+0.558$, $-0.554$,
and $+0.568^\circ$/mm after their respective response transitions.

Across 11 non-overlapping 10-mm windows within
these regimes, the linear fitting errors were
only $0.032$--$0.200^\circ$ RMSE.
Furthermore, fitting the first 60\% of each
window and predicting the remaining 40\%
yielded RMSEs of $0.035$--$0.388^\circ$.
These results demonstrate that local response
variations can be captured effectively by
first-order approximations.

\begin{figure*}[!t]
    \centering
    \includegraphics[width=0.98\textwidth]
    {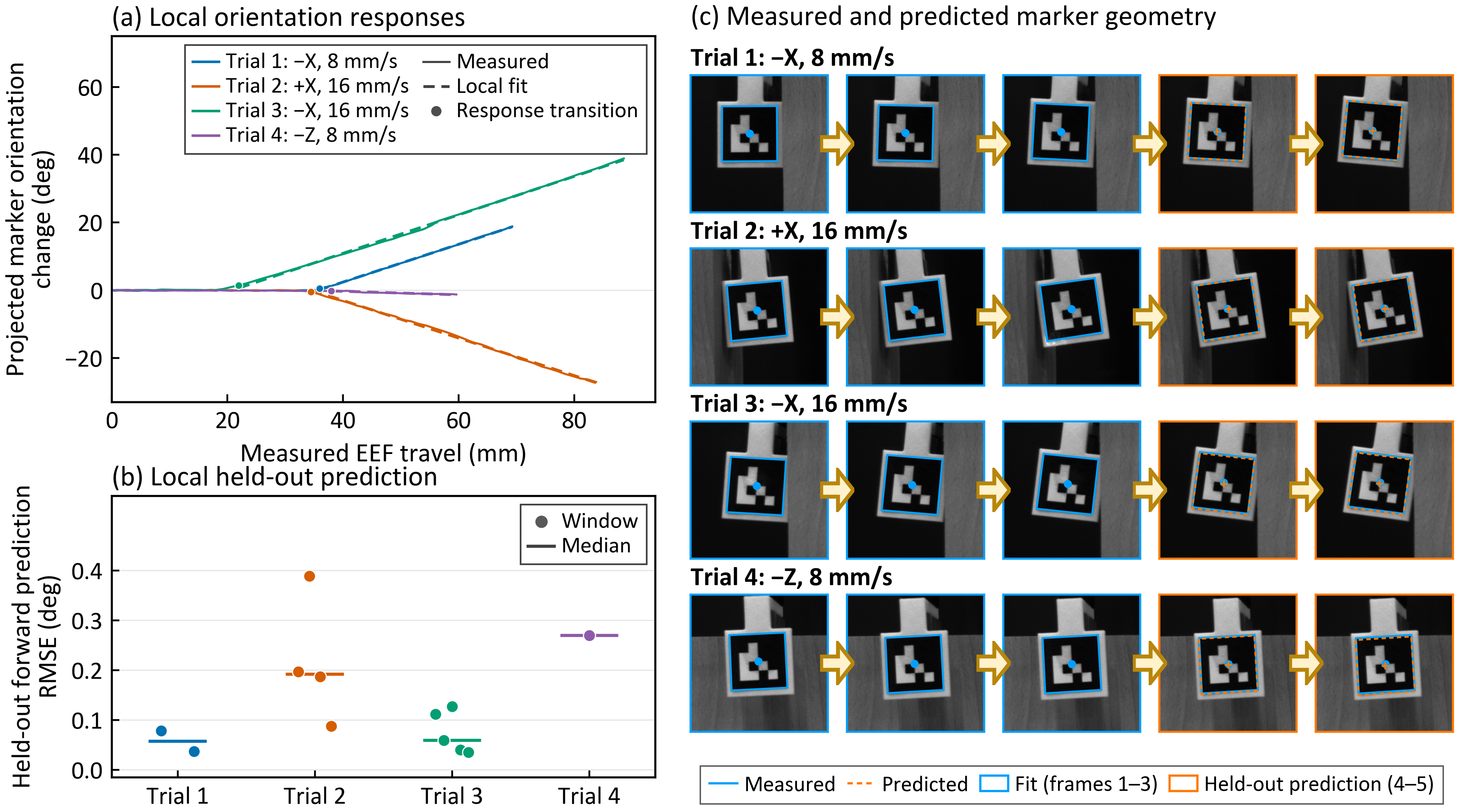}
    \caption{Experimental validation of local
    action--response linearity under external contact.
    (a) ArUco image-plane orientation changes
    with measured end-effector travel in four
    representative executions. Solid curves denote
    measurements, dashed lines indicate local linear
    fits, and markers indicate fitted response
    transitions.
    (b) Held-out forward-prediction errors across
    12 local motion windows from the four executions.
    Each local model is fitted using the first 60\%
    of observations and evaluated on the remaining
    40\%. Horizontal lines indicate median errors.
    (c) Representative video-frame sequences showing
    the measured and predicted ArUco marker geometry.
    Arrows indicate temporal progression, with blue
    and orange outlines distinguishing measured and
    predicted marker geometry during the held-out
    prediction stage.}
    \label{fig::local_linearity_validation}
\end{figure*}

As shown in Fig.~\ref{fig::local_linearity_validation},
the local approximation remains accurate over
limited motion ranges, whereas prediction errors
increase when the same model is extrapolated
across substantially different contact responses.
For example, a model fitted to the initial
10~mm of one trajectory yielded a prediction
RMSE of $0.034^\circ$ over 20~mm, compared
with $5.842^\circ$ when extended to 60~mm.

These observations provide experimental evidence
for the local linearity exploited in Section~V-B.
In particular, the results support evaluating
actions using locally estimated response
sensitivities instead of constructing a single
global contact-dynamics model.

\section{Discussion}

This paper presents a real-time, closed-loop framework for in-hand object reorientation, built upon the principles of perceptual action evaluation and adaptive action adjustment. By introducing a contact-agnostic evaluation metric based on task-relevant visual and tactile expectations, along with a local exploration strategy for action adjustment, the proposed method enables dynamic adaptation to diverse and unpredictable contact conditions. Experimental results demonstrate that the framework consistently outperforms conventional approaches, achieving higher success rates even in the presence of external disturbances and nonlinear contact dynamics. Notably, the method generalizes effectively across novel objects and environments without requiring prior contact models, offline training, or manual annotations—highlighting its adaptability and applicability to real-world robotic manipulation tasks. This adaptability is crucial for enhancing dexterity under limited actuation and uncertain contacts, and also lays the groundwork for future strategies that may actively leverage environmental forces.

Central to this capability is the task-oriented decomposition of the reorientation problem and the use of finger-object contact states as real-time indicators of action effectiveness. This design serves two key purposes: it filters out the confounding influence of external disturbances on action evaluation, and it enables a compact, task-relevant action space. Together, these features promote more efficient action exploration and fine-grained adjustment, ultimately contributing to the adaptability and responsiveness of the overall system.

Beyond the task of object reorientation, the underlying principles of this framework provide a generalizable foundation for a broader class of contact-rich manipulation tasks—such as cutting and insertion—where uncertainty, limited prior knowledge, and external perturbations are prevalent. Moving forward, we aim to extend these strategies to active environment exploitation, contributing to the development of robotic systems that are truly autonomous in complex, unstructured environments.

\bibliographystyle{IEEEtran}
\bibliography{ref}

\clearpage
\appendices

\setcounter{topnumber}{1}
\makeatletter
\setlength{\@fptop}{0pt}
\makeatother

\section{PPO Baseline: Implementation and Evaluation Protocol}
\label{app:ppo}

This appendix distinguishes the retained PPO checkpoint from the archived
environment and deployment source. Checkpoint parameters and the associated
training log agree; however, the available training script defines a different
network revision. Accordingly, environment details below describe the archived
implementation, not a verified reconstruction of every setting used to produce
the checkpoint. The exact robot adapter and a checkpoint identifier for each
reported trial were not preserved.

\subsection{Simulation Environment and Interfaces}
The archived MuJoCo environment uses a two-finger gripper and a supported object.
Its integration step is $0.5$\,ms, with gravity $[0,0,-9.81]$\,m/s$^2$.
Each policy action invokes a position controller and a separate force controller
with a fixed target of $20$\,N. The controller executes at most 1000 simulation
steps and can stop earlier after position, velocity, and force errors fall below
their tolerances. Thus, an action lasts at most $0.5$\,s of simulated time;
this is not a fixed policy sampling period.

Each finger provides a $20\times20$ three-channel contact-force grid. At each
taxel, the normal channel is set to zero and the two tangential channels are
divided by their Euclidean norm (zero vectors remain zero). Let
$\bar{\mathbf u}^{R}_t,\bar{\mathbf u}^{L}_t\in\mathbb R^3$ be the means over
taxels whose resulting norms exceed $0.1$ and $0.01$, respectively, with a
zero mean for an empty set. The policy input is
\begin{equation}
 \mathbf o_t=
 [\bar{\mathbf u}^{R\top}_t,\bar{\mathbf u}^{L\top}_t,
   \phi_t,\theta_t,\psi_t]^\top\in\mathbb R^9 .
 \label{eq:ppo-observation}
\end{equation}
Components 1 and 4 are identically zero. The last three components are object
roll, pitch, and yaw in radians, computed directly from the simulator quaternion;
they are privileged state information. Object position, images, joint states,
and normal-force magnitude are not policy inputs. The declared observation box
is $[-2,2]^9$, but the source does not clip the observations or normalize them
with running statistics; Euler angles can exceed these declared bounds.

During training, a Gaussian sample $\tilde{\mathbf a}_t$ is clipped by the
framework before environment execution:
\begin{equation}
 \mathbf a_t=\operatorname{clip}(\tilde{\mathbf a}_t,-1,1),
 \qquad [\Delta y_t,\Delta z_t]^\top=0.0045\,\mathbf a_t\ \mathrm{m}.
 \label{eq:ppo-action}
\end{equation}
There is no learned grasp-force or rotational action. An attempted minimum-action
adjustment in the source occurs after the scaled command is formed and therefore
does not affect the executed command.

\subsection{Reward and Episode Termination}
Let $\mathbf h_t=(y_{\rm hand},z_{\rm hand}+0.13\,\mathrm{m})$ and
$\mathbf p_t=(y_{\rm obj},z_{\rm obj})$ denote planar positions. With distances
expressed in meters, define
\begin{equation}
 D_t=\|\mathbf h_t-\mathbf p_t\|_2,\quad
 s_t=1000|D_t-D_0|,\quad e_t=|\phi_t|\frac{180}{3.14159}.
 \label{eq:ppo-errors}
\end{equation}
Here $D_0$ is measured after reset, $s_t$ is a radial-distance-change proxy in
millimeters, and $e_t$ is absolute roll error in degrees relative to zero.
The executed source reward is
\begin{equation}
\begin{split}
 r_t={}&-3s_t-0.2e_t
       +100\,\mathbf 1[e_t<5\ \land\ s_t<20]\\
      &-100\,\mathbf 1[s_t>20]-100\,\mathbf 1[k_t>100],
\end{split}
\label{eq:ppo-reward}
\end{equation}
where $k_t$ counts actions since reset, including the current action. Each
indicator is evaluated independently, so terminal contributions can accumulate.
Any of the three conditions terminates the episode; the timeout first occurs
at action 101. The code returns \texttt{terminated=True} for this timeout and
always returns \texttt{truncated=False}. There is no additional contact reward,
orientation-progress difference, reward clipping, or reward normalization.

\subsection{Policy Architecture and Optimization}
The retained artifact uses Stable-Baselines3 (SB3) 2.4.1
\cite{ppoapp_raffin2021stable}. Its actor and critic have separate learned
parameters. Table~\ref{tab:ppo-implementation} includes the additional linear
actor layer found in the weights and serialized policy definition; there is
no tanh after that layer. The action distribution is a diagonal Gaussian with
two learned, state-independent log standard deviations. In contrast, the
available training script specifies $[128,64]$ branches and an additional tanh;
that script is not the exact checkpoint-producing revision.

For the standard clipped PPO objective~\cite{ppoapp_schulman2017ppo}, let
$\rho_t(\vartheta)=\pi_\vartheta(\tilde{\mathbf a}_t\mid\mathbf o_t)/
\pi_{\rm old}(\tilde{\mathbf a}_t\mid\mathbf o_t)$ and let $\hat A_t$ denote
the minibatch-normalized generalized advantage estimate. SB3 minimizes
\begin{equation}
\begin{split}
 \mathcal L={}&-\mathbb E_t\!\left[\min\!\left(
 \rho_t\hat A_t,\operatorname{clip}(\rho_t,0.9,1.1)\hat A_t\right)\right]\\
 &+0.5\,\mathbb E_t[(V_\vartheta(\mathbf o_t)-\hat R_t)^2]
 -0.02\,\mathbb E_t[\mathcal H(\pi_\vartheta(\cdot\mid\mathbf o_t))].
\end{split}
\label{eq:ppo-objective}
\end{equation}
The checkpoint records 115,200 training transitions, or 30 rollouts of
$30\times128$ transitions, and 450 optimization epochs. The learning rate
decreases linearly from $3\times10^{-4}$ to $10^{-5}$ during the first 90\%
of training, then remains constant. The training log confirms this schedule
and evaluations every 3840 transitions. Per-object episode counts, training
seeds, and the rule selecting the checkpoint used on the robot cannot be
established from these records; in particular, the retained artifact is not
verified as the best validation checkpoint.

\begin{table}[t]
\caption{Parameters verified in the retained PPO checkpoint.
R and L denote ReLU and linear outputs, respectively.}
\label{tab:ppo-implementation}
\centering
\small
\setlength{\tabcolsep}{3pt}
\begin{tabular}{@{}>{\raggedright\arraybackslash}p{0.36\columnwidth}>{\raggedright\arraybackslash}p{0.60\columnwidth}@{}}
\toprule
Parameter & Value \\
\midrule
Input / output dimension & 9 / 2 \\
Actor mean network & $9\!\to\!64_{\rm R}\!\to\!64_{\rm R}\!\to\!64_{\rm L}\!\to\!2_{\rm L}$ \\
Critic network & $9\!\to\!64_{\rm R}\!\to\!64_{\rm R}\!\to\!1_{\rm L}$ \\
Transitions / parallel envs. & 115,200 / 30 \\
Rollout / batch size & 128 per environment / 128 \\
Epochs per rollout & 15 \\
Discount / GAE & $\gamma=0.99$, $\lambda=0.95$ \\
Policy clip / gradient cap & 0.1 / 0.5 (global norm) \\
Value / entropy weight & 0.5 / 0.02 \\
Optimizer & Adam; $\beta=(0.9,0.999)$, $\epsilon=10^{-5}$ \\
Advantage normalization & Per minibatch; no value clipping \\
\bottomrule
\end{tabular}
\end{table}

\subsection{Reset-Time Condition Variation}
The archived reset samples $c\sim\operatorname{Uniform}\{0.5,1.0,1.5\}$ and
sets the floor's sliding, torsional, and rolling friction parameters to
$(c,c,c)$. These are randomized friction parameters, not three fixed evaluation
conditions or a single effective contact coefficient. A scripted grasp sequence
also samples a vertical command uniformly from
$[z_o-0.13+0.5|y_o|,\,2|y_o|+0.02]$ meters, using object coordinates at reset.
Object selection samples the enabled XML list, but the retained default list
contains only the wrench. Separate counters name seven objects without a
consistent link to the retained run. Consequently, neither its exact object
mixture nor its executed randomization configuration is established.
There is no explicit sampling of mass, center of mass, geometry scale, gravity,
sensor/action noise, or external disturbances in the archived reset.

\begin{algorithm}[t]
\caption{Recovered PPO training and evaluation protocol}
\label{alg:ppo-protocol}
\small
\DontPrintSemicolon
\textbf{Source-defined environment:} reset the object, sample floor friction
and vertical grasp offset, execute the grasp sequence, and record $D_0$\;
\textbf{Checkpoint-confirmed optimization:} use 30 environments and the
policy and optimizer in Table~\ref{tab:ppo-implementation}\;
\While{fewer than 115,200 training transitions have been collected}{
 Collect 128 transitions per environment: construct
 \eqref{eq:ppo-observation}, sample and clip the action, execute the controller,
 compute \eqref{eq:ppo-reward}, and reset terminated episodes\;
 Apply SB3 GAE and 15 PPO epochs with minibatches of 128 using
 \eqref{eq:ppo-objective} and the recorded learning-rate schedule\;
}
Retain the saved policy; a best-checkpoint selection rule is unverified\;
\textbf{Robot interface:} the archived loader requests deterministic prediction
from the saved policy, with an external robot controller and no online
\texttt{learn} call; the executed adapter revision is unavailable\;
Score the recorded robot trials using the criterion in Sec.~VI-B,
separately from simulation termination\;
\end{algorithm}

\subsection{Deployment, Evaluation, and Comparison Scope}
The deployment source loads the saved PPO model and requests
deterministic prediction, but contains unfinished observation/action mappings
and is not an executable record of the historical adapter. Sampled robot traces
contain nine observation components: a normalized tactile cue with a mirrored
counterpart and a single measured angle in radians, with two remaining angles
zero. This differs from two simulated force grids and simulator Euler angles.
The precise tactile transformation, coordinate calibration, control period,
force logic, and online stopping gates therefore remain unverified.

The reported robot success criterion is an orientation error below $5^\circ$
with displacement within 30\,mm, rather than the simulator's 20\,mm condition.
Sampled robot traces also show a $-100$ reward contribution above 30\,mm and
continued execution after that contribution; it cannot be equated with
immediate simulation termination. Table~II is retained as recorded in the
experiment workbook. Its normalized-step means use available step entries;
for the toy, seven PPO step entries are available for eight recorded successes.

The comparison evaluates the recorded end-to-end systems. PPO has two learned
translation outputs and privileged pose during simulation. The archived
proposed controller instead selects and adjusts discrete translation and
grasp-force primitives online, with tactile gating of force adjustment and
separate visual and tactile reward terms. These interface differences, the
unverified deployment revision, and incomplete run-to-checkpoint provenance
limit attribution of the measured performance gap to the learning algorithm
alone.

\end{document}